\documentclass{article}

\usepackage[preprint, nonatbib]{neurips_2024}

\usepackage[utf8]{inputenc} 
\usepackage[T1]{fontenc}    
\usepackage{hyperref}       
\usepackage{url}            
\usepackage{booktabs}       
\usepackage{amsfonts}       
\usepackage{nicefrac}       
\usepackage{microtype}      
\usepackage{xcolor}         

\usepackage{natbib}
\setcitestyle{authoryear,open={[},close={]}}

\usepackage{url}
\usepackage{bm}
\usepackage{graphicx}
\usepackage{commath}
\usepackage{wrapfig}

\usepackage{multicol}
\usepackage{multirow}
\usepackage{algorithm}
\usepackage{algpseudocode}
\usepackage{subfigure}

\usepackage{xcolor}
\usepackage{colortbl}
\usepackage{textcomp}
\definecolor{my-color}{rgb}{0.28, 0.58, 0.28}
\definecolor{Gray}{gray}{0.85}

\usepackage{mathtools}
\DeclarePairedDelimiter{\nint}\lfloor\rceil

\title{ApiQ: Finetuning of 2-Bit \\
    Quantized Large Language Model}

\author{
  Baohao Liao$^{1,2}$\thanks{Correspondence to: \texttt{\href{mailto:b.liao@uva.nl}{b.liao@uva.nl}}} \:\:\:\:\: Christian Herold$^{2}$ \:\:\:\:\: Shahram Khadivi$^{2}$ \:\:\:\:\: Christof Monz$^{1}$ \\
  $^{1}$Language Technology Lab, University of Amsterdam \\
  $^{2}$eBay Inc., Aachen, Germany \\
  Code: \href{https://github.com/BaohaoLiao/ApiQ}{https://github.com/BaohaoLiao/ApiQ}
}

\begin{document}

\maketitle

\begin{abstract}
Memory-efficient finetuning of large language models (LLMs) has recently attracted huge attention with the increasing size of LLMs, primarily due to the constraints posed by GPU memory limitations and the effectiveness of these methods compared to full finetuning. Despite the advancements, current strategies for memory-efficient finetuning, such as QLoRA, exhibit inconsistent performance across diverse bit-width quantizations and multifaceted tasks. This inconsistency largely stems from the detrimental impact of the quantization process on preserved knowledge, leading to catastrophic forgetting and undermining the utilization of pretrained models for finetuning purposes. In this work, we introduce a novel quantization framework, \textit{ApiQ}, designed to restore the lost information from quantization by concurrently initializing the LoRA components and quantizing the weights of LLMs. This approach ensures the maintenance of the original LLM's activation precision while mitigating the error propagation from shallower into deeper layers. Through comprehensive evaluations conducted on a spectrum of language tasks with various LLMs, ApiQ demonstrably minimizes activation error during quantization. Consequently, it consistently achieves superior finetuning results across various bit-widths.
\end{abstract}

\begin{figure}[ht]
  \centering
  \includegraphics[width=0.98\textwidth]{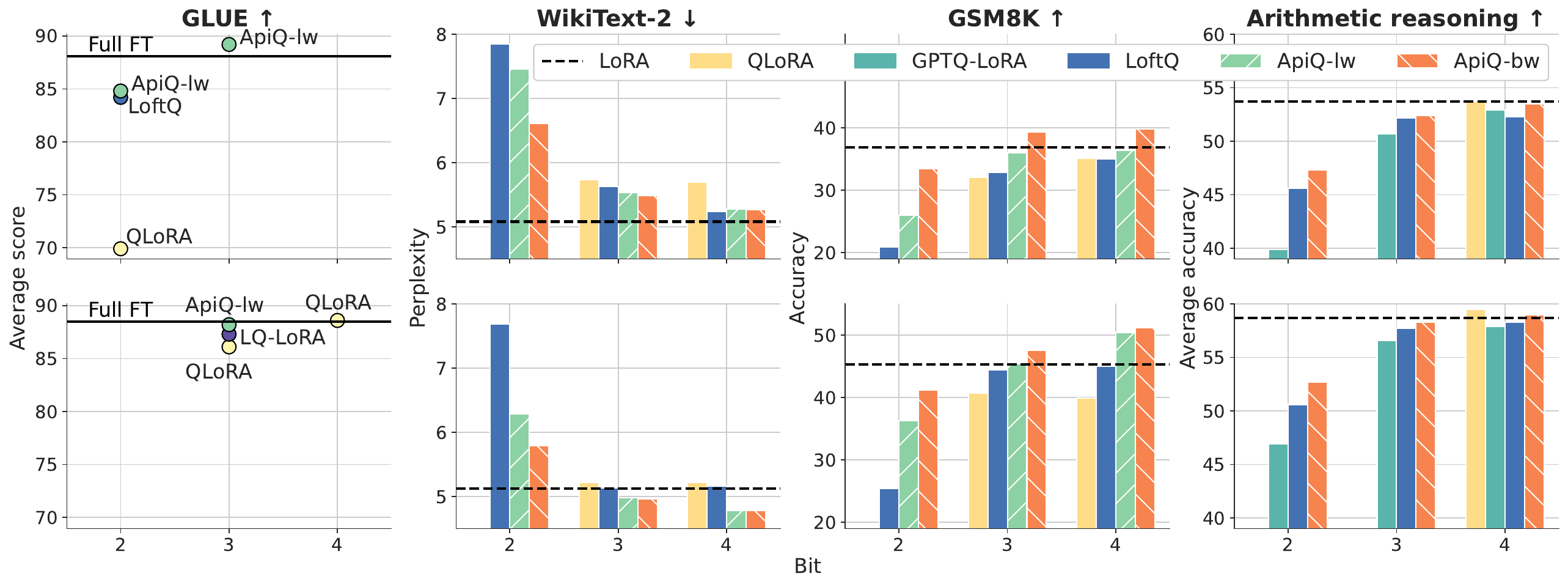}
  \caption[]{Finetuning performance over various tasks. \textbf{1st row}: LLM is DeBERTa-v3-base for GLUE and Llama-2-7B for the rest. \textbf{2nd row}: LLM is RoBERTa-large for GLUE and Llama-2-13B for the rest. For better visualization, some extremely worse results from (2-bit or 3-bit) QLoRA are ignored.}
  \label{fig: overall}
\end{figure}

\section{Introduction}
\label{sec: introduction}
Large language models (LLMs) have garnered significant acclaim and success across a wide range of domains and applications \citep{llama-2, mistral, gpt4}. With ongoing advancements, the scope and complexity of released LLMs have witnessed exponential growth, with some LLMs encompassing more than 50B parameters \citep{llama-2, llama, opt, bloom}. This remarkable upscaling introduces considerable challenges, particularly when effectively adapting these models for downstream tasks. Historically, a prevalent method for adapting pretrained models to specific downstream tasks is full finetuning, a process that updates all pretrained parameters. Although this approach has led to many state-of-the-art achievements, its practicality is somewhat hindered in scenarios where GPU memory is limited. This limitation stems from the necessity to store the model's weights and optimizer's states in the GPU's memory and intensifies with the escalating sizes of LLMs.

To mitigate the extensive memory requirement for finetuning LLMs, an alternative method is parameter-efficient finetuning (PEFT) \citep{hiwi, lora, adapter}. This technique involves selectively introducing and updating a limited set of parameters while leaving the majority of the LLM's parameters unchanged. A key advantage of this approach is the substantial reduction in GPU memory required for the optimizer's states since the size of the optimizer states is proportional to the amount of trainable parameters. To further reduce the GPU memory required by loading the LLM's weights, multiple model compression methods have been proposed \citep{omniquant, smoothquant, spqr, awq, gptq}, converting high-precision weight values into a discrete set of values. Initially, quantization techniques were primarily developed for deploying LLMs in memory-limited environments for inference purposes. QLoRA \citep{qlora} innovatively combines PEFT, specifically LoRA \citep{lora}, with quantization methods to remarkably reduce the GPU memory requirement for finetuning.

However, QLoRA inherits the same challenge as LLM's quantization, namely the distortion of the learned knowledge from the full-precision LLM due to the quantization error, which exacerbates at lower-bit quantizations, leading to catastrophic forgetting. Recently, \citet{loftq} and \citet{lq-lora} proposed a new method to reduce the quantization error through a strategic initialization of the LoRA components to maintain the original weight states. This technique has demonstrated considerable success in the finetuning of lower-bit quantized LLMs (QLLMs). Nonetheless, they focus on preserving the weight states on a per-linear-layer basis, resulting in accumulative error propagation through layers.

In this paper, we introduce a novel and efficient quantization framework, \underline{a}ctivation-\underline{p}reserved \underline{i}nitialization of \underline{Q}LLM termed \textit{ApiQ}, which consists of two steps to adapt an LLM, similar to QLoRA. During the quantization step, ApiQ preserves the activation instead of the weight of full-precision LLM by jointly optimizing the LoRA's components and quantizing the LLM's weights. This approach ensures that the output of the QLLM remains consistent with that of the full-precision LLM, effectively mitigating quantization error by aligning the activations across corresponding layers. As a result, the quantization errors introduced in earlier layers are ameliorated. Subsequently, we finetune the LoRA modules with the fixed QLLM on downstream tasks, thereby significantly reducing the demands on GPU memory.

Our primary contributions are as follows:
\begin{itemize}
\item We conduct an in-depth analysis of the challenges associated with finetuning QLLM (\S\ref{sec: challenges}).
\item We propose ApiQ to initialize the PEFT parameters in conjunction with the quantization of an LLM, aiming to retain the activation of the full-precision LLM (\S\ref{sec: method}).
\item ApiQ demonstrates superior performance post-quantization, even surpassing the latest post-training quantization (PTQ) techniques (\S\ref{sec: quantization quality}).
\item We carry out extensive finetuning experiments on 5 LLMs across 5 different tasks to evaluate the effectiveness of ApiQ. ApiQ consistently outperforms all baselines at various bit levels (\S\ref{sec: finetuning results} and Figure \ref{fig: overall}).
\end{itemize}

\section{Preliminaries}
\label{sec: preliminaries}
\begin{wrapfigure}{r}{0.4\textwidth}
    \includegraphics[width=\linewidth]{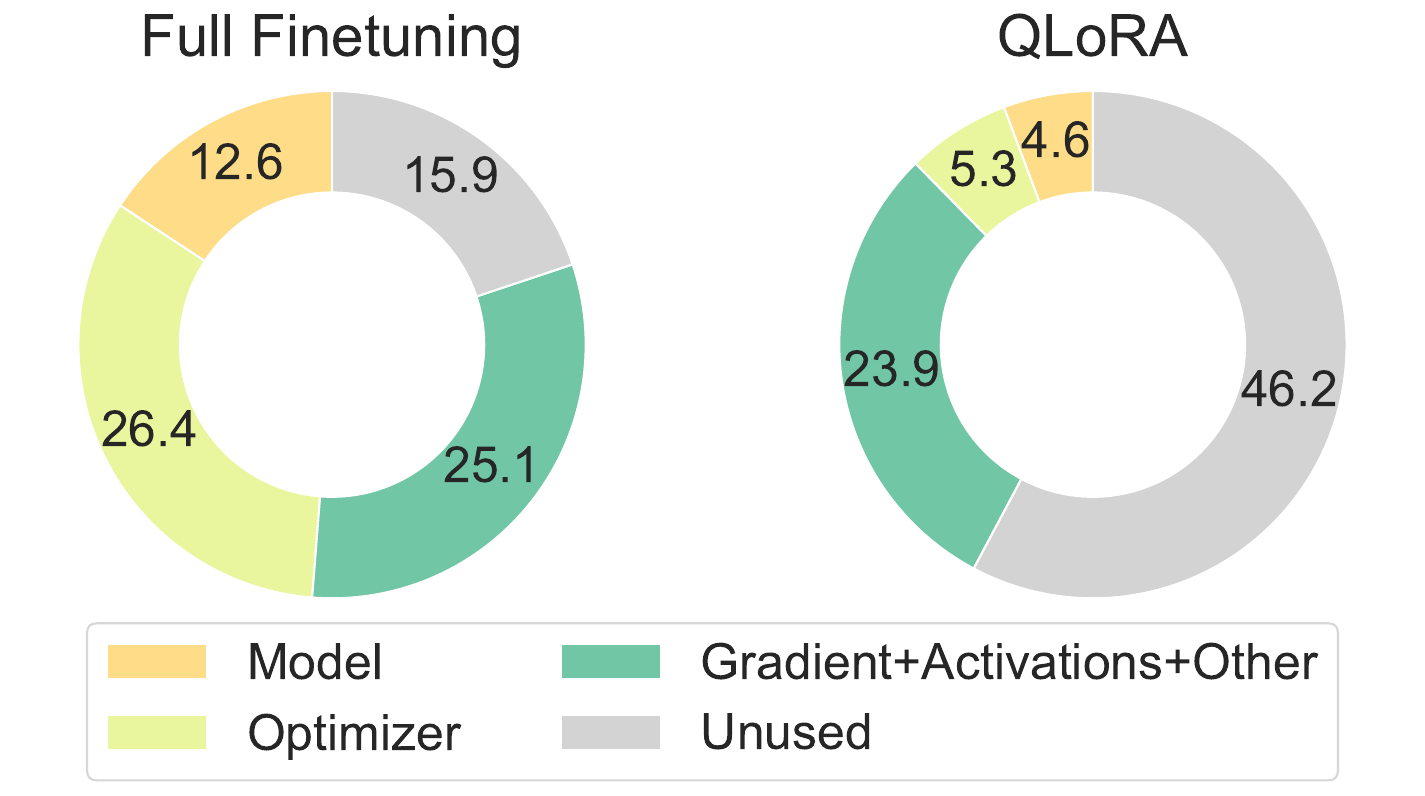} 
    \caption[]{Memory allocation (GB) of a A100-80GB GPU for finetuning Llama-2-7B. The optimizer is Adam. The batch size is 1. The sequence length is 2048. For QLoRA, the bit-width is 4 and the LoRA rank is 64.}
    \label{fig: mem}
\end{wrapfigure}
\textbf{GPU memory allocation} and utilization are typified by three principal mechanisms for training a model, as exemplified in Figure \ref{fig: mem} during the finetuning of Llama-2-7B \citep{llama-2}. In full finetuning, a substantial portion of GPU memory is allocated to store the model's weights. For instance, approximately 12.6GB is required for a model comprising roughly 7B parameters in BFloat16 format. Secondly, the optimizer states associated with trainable parameters occupy a considerable amount of GPU memory. Employing Adam \citep{KingmaB14} as the optimizer necessitates storing the first and second moments in the GPU memory, effectively doubling the memory requirement compared to that needed for the trainable parameters alone. Notably, the memory allocations for the model's weights and optimizer states are static, remaining constant throughout the training process. The third aspect involves the temporary storage of \textit{activations} — the outputs produced by each layer as data traverses through the model. These activations are crucial for gradient computation during the backward pass and are retained in memory for this purpose. After the gradient computation, these activations are discarded. Modern training frameworks, like PyTorch \citep{pytorch}, employ a sequential process for gradient computation and activation deletion, enhancing memory efficiency. Subsequently, gradients are utilized to update the model's weights and optimizer states, and then they too are eliminated. Peak memory usage typically occurs at the onset of gradient computation or during the update of optimizer states.

\textbf{Memory-efficient finetuning.} In response to the GPU memory constraints and the increasing size of LLMs, various strategies have been developed to optimize memory efficiency during finetuning. To mitigate activation memory demands, techniques such as selective activation storage and on-demand recomputation are employed \citep{meft, revnet, gradientCheckpointing}. Additionally, to curtail the memory required for optimizer state storage, the pretrained LLM is kept fixed while a limited amount of trainable parameters are introduced \citep{lora, adapter}. A prime example is LoRA \citep{lora}, which adapts the pretrained weight, $\bm{W}\in\mathbb{R}^{d_1 \times d_2}$, of a linear layer as $\bm{W}' = \bm{W} + \frac{\alpha}{r}\bm{AB}^\top$, where $\bm{A}\in\mathbb{R}^{d_1 \times r}$, $\bm{B}\in\mathbb{R}^{d_2 \times r}$, $r \ll d_1$, $r \ll d_2$ and $\alpha$ is a scalar. Introducing a smaller bottleneck dimension $r$ substantially reduces the memory demand for the optimizer, illustrated by a reduction from 26.4GB to 5.3GB as shown in Figure \ref{fig: mem}. To further reduce LoRA’s memory usage, \citet{qlora} introduced a quantized version of $\bm{W}$, such as a 4-bit representation in contrast to 16 bits. This technique significantly decreases the memory requirement for storing the model's weights, from 12.6GB to 4.6GB.

\textbf{Quantization} involves converting high-precision values into discrete levels. In this research, we focus on uniform affine quantization \citep{uniformQuant}, known for its enhanced hardware support and efficiency. This process quantizes the pretrained weight as follows:
\begin{align}
    \bm{W}_q = \mathrm{clamp}(\nint{\frac{\bm{W}}{s}} + z, 0, 2^b-1) \label{eq: quantization}
\end{align}
where the scale factor $s=\frac{\mathrm{max}(\bm{W})-\mathrm{min}(\bm{W})}{2^b-1}$, the zero-point $z = -\nint{\frac{\mathrm{min}(\bm{W})}{s}}$, $b$ is the bit-width, and $\nint{}$ is the round-to-nearest operation. One only needs to load $\bm{W}_q$ and $z$ in a reduced bit format, and $s$ in Float16 to GPU. During the forward pass, they are de-quantized for activation computation as $\bm{Q} = s(\bm{W}_q - z)$.

\section{Challenges of finetuning QLLM}
\label{sec: challenges}
QLoRA \citep{qlora} employs a strategy wherein the fixed pretrained weights are loaded onto the GPU in a lower-bit format, while finetuning is confined to a minimal number of parameters from the adapters. This approach significantly reduces the memory allocation required from both the model's weights and optimizer states, decreasing it from 39GB to 10GB, as depicted in Figure \ref{fig: mem}. This reduction in memory demand facilitates the finetuning of LLMs on more modest computational resources. Nevertheless, this method introduces certain challenges associated with quantization.

\begin{figure}[t]
  \centering
  \includegraphics[width=0.98\linewidth]{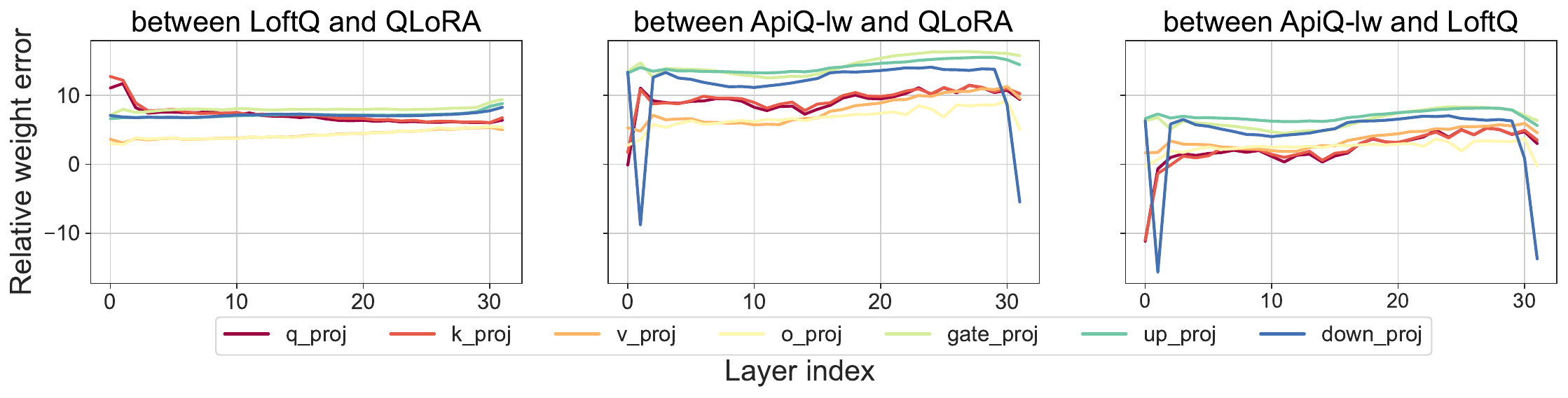}
  \caption{Relative weight quantization error of 2-bit quantized Llama-2-7B, i.e. \(e = \lVert\delta W^{\text{baseline}}\rVert_F - \lVert\delta W^{\text{method}} \rVert_F\). The larger $e$ is, the more effective the method is in reducing weight error compared to the baseline. \textbf{Left}: The method is LoftQ and the baseline is QLoRA. \textbf{Middle}: The method is ApiQ and the baseline is QLoRA. \textbf{Right}: The method is ApiQ and the baseline is LoftQ. Refer to Figure \ref{fig: weight diff} for the 2-bit and 4-bit non-relative weight error.}
  \label{fig: relative weight diff}
\end{figure}
\subsection{QLLM breaks down the starting point}
\label{sec: starting point}
LLMs are recognized for their ability to learn broadly applicable and distributed representations that effectively support the downstream learning of compressed task-specific representations \citep{DBLP:conf/acl/AghajanyanGZ20}, i.e. offering a robust starting point for the training of downstream tasks. \citet{meft} postulate that maintaining this starting point — ensuring that the difference between the modified weight $\bm{W}'$ and the original weight $\bm{W}$ is minimal (i.e., $\lVert\bm{W}'-\bm{W}\rVert\rightarrow0$) — is crucial at the finetuning's outset to achieve performance comparable to full finetuning. 

LoRA \citep{lora} adheres to this principle by initializing $\bm{B}=\bm{0}$, which results in $\bm{W}'=\bm{W}$ at the start of the training. QLoRA, on the other hand, follows LoRA's default initialization for $\bm{A}$ and $\bm{B}$. Consequently, at the onset of training, $\bm{W}' = \bm{Q} + \bm{AB^\top} = \bm{Q}$. Due to the round-to-nearest and clipping operations involved in quantization, $\bm{Q}$ differs from the original $\bm{W}$, thereby distorting the starting point. This deviation, represented by $\lVert\delta\bm{W}\rVert = \lVert\bm{W}-\bm{W}'\rVert$, is expected to increase with lower-bit quantization.

Recent developments by \citet{loftq} and \citet{lq-lora} introduced an approach to initialize the $\bm{Q}$, $\bm{A}$ and $\bm{B}$ matrices in QLoRA by solving the following optimization problem:
\begin{align}
\underset{\bm{Q}, \bm{A}, \bm{B}}{\mathrm{argmin}} \; \lVert\bm{W} - (\bm{Q} + \bm{AB}^\top)\rVert \label{eq: loftq}
\end{align}
The key objective of this technique is to obtain $\bm{Q}$, $\bm{A}$, and $\bm{B}$ in such a way that the initial state of the model (the starting point) is preserved as closely as possible. As shown in Figure \ref{fig: relative weight diff} (Left), LoftQ \citep{loftq} significantly reduces the weight error.

\begin{table}[t]
  \centering
  \scriptsize
  \caption{The effect of trainable LoRA position. All linear layers are incorporated with a LoRA module initialized with different methods. Only the LoRA modules in the denoted position are finetuned. ApiQ has the smallest gap between different positions. \label{tab: partial lora}}
  \begin{tabular}{ll|ccc}
  \toprule
  \multirow{2}{*}{\textbf{Method}} & \textbf{LoRA} & \textbf{MNLI} (acc$\uparrow$) &  \multicolumn{2}{c}{\textbf{WikiText-2} (ppl$\downarrow$)} \\
   & \textbf{position} & \textbf{2 Bits} & \textbf{4 Bits} & \textbf{2 Bits} \\
  \midrule
  \multirow{3}{*}{QLoRA} & All & \textbf{79.7} & \textbf{5.24} & N.A. \\
  & FFN & 78.2 & 5.29 & N.A. \\
  & Attn & 75.7 & 5.28 & N.A. \\
  \midrule
  \multirow{3}{*}{LoftQ} & All & \textbf{88.5} & \textbf{5.24} & \textbf{7.85} \\
  & FFN & 87.1 & 5.30 & 8.64 \\
  & Attn & 87.5 & 5.28 & 8.86 \\
  \midrule
  \multirow{3}{*}{ApiQ-lw} & All & \textbf{88.6} & 5.28 & \textbf{7.46} \\
  & FFN & 88.2 & 5.29 & 7.50 \\
  & Attn & \textbf{88.6} & \textbf{5.25} & 7.55 \\
  \bottomrule
  \end{tabular}
  \centering
\end{table}
\subsection{Accumulative quantization error}
\label{sec: error accumulation}
The findings of \citet{lora} highlight that integrating LoRA modules solely into the query and value projection layers is adequate to match the performance of full finetuning. This stands in contrast to the stance of \citet{qlora}, who advocate for integrating the LoRA modules into all linear layers of QLLM to achieve similar results.

We extend upon the conclusion of QLoRA by conducting finetuning experiments with DeBERTa-v3-base \cite{deberta} and Llama-2-7B \cite{llama-2} on the MNLI \citep{mnli} and WikiText-2 \citep{wikitext} datasets, respectively. As presented in Table \ref{tab: partial lora}, the most effective results from QLoRA are achieved when the LoRA modules in all linear layers are trained, rather than a subset of them. This observation leads us to propose that each linear layer undergoes a loss of learned information as a consequence of quantization. To mitigate this loss and restore the original learned information, it is essential to adapt each linear layer individually with a LoRA module.

Furthermore, we notice that the quantization errors accumulate through layers. Consider two consecutive linear layers, $\bm{W}_0$ and $\bm{W}_1$, with inputs and outputs $\bm{X}_0$, $\bm{X}_1$ and $\bm{X}_2$, respectively. Under QLoRA's quantization, the activation error for the first layer is $\lVert\bm{X}_1 - \bm{X}_1^{q}\rVert = \lVert\bm{X}_0\bm{W}_0 - \bm{X}_0\bm{Q}_0\rVert=\lVert\bm{X}_0\bm{W}_0 - \bm{X}_0(\bm{W}_0 - \delta\bm{W}_0)\rVert = \lVert\bm{X}_0\delta\bm{W}_0\rVert$, where $\delta\bm{W}_0$ is the quantization error. For the second layer, the error is $\lVert\bm{X}_2 - \bm{X}_2^{q}\rVert = \lVert\bm{X}_0\bm{W}_0\delta\bm{W}_1+\bm{X}_0\delta\bm{W}_0\bm{W}_1 - \bm{X}_0\delta\bm{W}_0\delta\bm{W}_1\rVert$. This phenomenon indicates that the quantization errors from shallower layers are propagated to deeper layers, with potentially greater impact in deeper LLMs. This effect underscores another justification for the implementation of LoRA modules in every quantized linear layer to timely counteract the errors.

Despite the advances made by LoftQ in reducing the quantization error $\delta\bm{W} = \bm{W} - (\bm{Q} + \bm{AB}^\top)$, the issue of error propagation persists. This is evidenced in Table \ref{tab: partial lora}, where the performance between training all LoRA modules and only training a subset of them is still large, especially for lower-bit quantization. Such findings emphasize the importance of not only minimizing the quantization errors at their source but also managing their propagation across layers.

\subsection{SVD is not a universal solution}
\label{sec: unique case}
\citet{loftq} and \citet{lq-lora} apply an iterative algorithm to solve Equation (\ref{eq: loftq}) as:
\begin{align}
    \begin{aligned}[c]
    \bm{A}^{(t)}, \bm{B}^{(t)} &\leftarrow \text{SVD}(\bm{W} - \bm{Q}^{(t-1)}) \nonumber \\
    \end{aligned}
    \qquad\qquad
    \begin{aligned}[c]
    \bm{Q}^{(t)} &\leftarrow  f(\bm{W} - \bm{A}^{(t)}\bm{B}^{{(t)}\top}) \nonumber 
    \end{aligned}
\end{align}
where $f$ is a function for a sequential quantization and de-quantization as:\footnote{We use uniform affine quantization to represent the quantization of LoftQ and LQ-LoRA for easy understanding. LoftQ and LQ-LoRA actually apply NF-quantization as QLoRA, causing slower inference speed than the uniform quantization.}
\begin{align}
\bm{Q} &= f(\bm{W}) = s \cdot (\mathrm{clamp}(\nint{\frac{\bm{W}}{s}} + z, 0, 2^b-1) - z) \label{eq: f}
\end{align}

Although this algorithm is effective without the usage of calibration data, we couldn't easily apply it to other PEFT methods, even a variant of LoRA, i.e. DoRA \citep{dora}. This algorithm requires a relationship of addition between the PEFT parameters and $\bm{W}$, which is not possible for some PEFT methods, like (IA)$^3$ \citep{ia3}, Adapter \citep{adapter}, HiWi \citep{hiwi} and so on.

Overall, to effectively finetune a QLLM, we need to preserve the starting point, mitigate the propagation of quantization error, and design a universal algorithm for various PEFT methods.

\section{Method: ApiQ}
\label{sec: method}
In this section, we introduce a novel quantization framework, \underline{A}ctivation-\underline{p}reserved \underline{i}nitialization of \underline{Q}LLM termed as \textit{ApiQ}, that addresses all above-mentioned challenges when finetuning a QLLM. The core objective of ApiQ is to maintain the integrity of the starting point, while effectively minimizing the cumulative impact of the quantization errors as they traverse through the network.

\subsection{Activation-preserved initialization}
The ApiQ framework introduces a distinct approach to quantization by focusing on minimizing the activation error, rather than the weight error as in previous methods \citep{loftq, lq-lora}. The core optimization problem of ApiQ is formulated as follows:
\begin{align}
\underset{\bm{Q}, \bm{A}, \bm{B}}{\mathrm{argmin}} \; ||\bm{X}\bm{W} - \bm{X}^q(\bm{Q} + \bm{AB}^T)|| \label{eq: apiq}
\end{align}
The pretrained weight $\bm{W} \in \mathbb{R}^{d_1 \times d_2}$ remains fixed during optimization. The quantized weight $\bm{Q} \in \mathbb{R}_b^{d_1 \times d_2}$ is represented in b-bit format, while $\bm{A} \in \mathbb{R}^{d_1 \times r}$ and $\bm{B} \in \mathbb{R}^{d_2 \times r}$ are low-rank matrices that are trainable. The input to the linear layer $\bm{W}$ is denoted as $\bm{X} \in \mathbb{R}^{n \times t \times d_1}$, where $n$ is the batch size and $t$ is the sequence length. Consequently, $\bm{X}\bm{W}$ represents the output or activation of the linear layer. The input to the quantized linear layer with LoRA is $\bm{X}^q \in \mathbb{R}^{n \times t \times d_1}$. It's important to note that for the first linear layer of an LLM, $\bm{X}$ equals $\bm{X}^q$. For subsequent layers, $\bm{X}^q$ is the output from the nearest shallower quantized layer of $\bm{W}$.

A key difference from LoftQ \citep{loftq} and LQ-LoRA \citep{lq-lora} is that ApiQ requires sequential optimization for each linear layer following the input order of different layers, as $\bm{X}_q$ is derived from the preceding layer. E.g., in each transformer block of Llama-2 \citep{llama-2}, the optimization should start with the key, query, and value projection layers, followed by the output projection layer, then the gate and up projection layer, and finally the down projection layer.

ApiQ has two primary advantages. Firstly, it ensures that the output from the quantized linear layer closely aligns with the original output, thereby preserving the starting point of the model. Secondly, it potentially mitigates the quantization error from shallower layers into deeper layers. This is achieved by consistently enforcing that the output of each quantized layer closely matches the original output, thereby gradually easing the quantization error as it propagates through the network. This mechanism isn't present in LoftQ and LQ-LoRA, giving ApiQ a unique advantage in managing and reducing the quantization errors in QLLMs.

\subsection{Block-wise ApiQ}
\label{sec: apiq-bw}
We define Equation (\ref{eq: apiq}) as \textit{layer-wise ApiQ} (\textit{ApiQ-lw}), because the LLM is quantized in a layer-by-layer manner. Additionally, we can optimize the entire transformer block simultaneously as:
\begin{align}
\underset{\bm{Q}s, \bm{A}s, \bm{B}s}{\mathrm{argmin}} \; ||\mathcal{F}(\bm{W}s, \bm{X}) - \mathcal{F}(\bm{Q}s, \bm{A}s, \bm{B}s, \bm{X}^q)|| \label{eq: apiq_bw} \nonumber
\end{align}
where $\mathcal{F}$ denotes the mapping function of a transformer block, $\bm{W}$s represent all the weights of the linear layers within this block, $\bm{X}$ is the input to this block, $\bm{Q}$s are the quantized versions of $\bm{W}$s, $\bm{A}$s and $\bm{B}$s are all low-rank matrices within this block, and $\bm{X}^q$ is the input to the quantized block and the output from the preceding quantized block. Block-wise ApiQ (\textit{ApiQ-bw}) necessitates sequential optimization but on a block-by-block basis, meaning we first optimize the first transformer block, followed by the second block, and so on.

ApiQ-bw retains the benefits of ApiQ-lw while offering two additional advantages. Firstly, ApiQ-bw is more time-efficient than ApiQ-lw because it quantizes the entire block in one step. Secondly, ApiQ-bw is compatible with a broader range of PEFT methods without necessitating adaptations for every linear layer. The matrices $\bm{A}$s and $\bm{B}$s do not have to be the low-rank matrices from LoRA; they can be trainable parameters from any PEFT method, such as DoRA, (IA)$^3$, HiWi and Adapter.

\begin{algorithm}[t]
   \caption{ApiQ-lw for one linear layer}
   \label{alg: apiq-lw}
   \footnotesize
   \begin{algorithmic}[1]
   \State {\bfseries Input:} calibration samples $\bm{X}$ and $\bm{X}^q$, $\bm{W}$, $\bm{A}$, $\bm{B}$, $\Theta$
   \State {\bfseries Output:} $\bm{Y}$, $\bm{Y}^q$, $\bm{A}$, $\bm{B}$, $\Theta$
   \State $\bm{Y} = \bm{X}\bm{W}$ \Comment{Unquantized output, save for next layer}
   \For{$e=0$ {\bfseries to} $\mathrm{(Epochs - 1)}$}
   \For{($\bm{y}, \bm{x}^q$) in ($\bm{Y}$, $\bm{X}^q$)} \Comment{Batch-wise}
   \State $\bm{Q} = \bm{W}/s$
   \State $\bm{Q} = \nint{\bm{Q}} + \bm{Q} - \bm{Q}.\mathrm{detach}()$  \Comment{Straight-through estimator}
   \State $\bm{Q} = s \cdot (\mathrm{clamp}(\bm{Q}+z, 0, 2^b-1) - z)$
   \State $\bm{y}^q = \bm{x}^q(\bm{Q}+\bm{AB}^\top)$ \Comment{Quantized output}
   \State $\mathrm{loss} = \mathrm{MSE}(\bm{y} - \bm{y}^q)$
   \State $\mathrm{loss.backward}()$
   \EndFor
   \EndFor
   \State $\bm{Y}^q = \bm{X}^q(\bm{Q}+\bm{AB}^\top)$ \Comment{Save for next layer}
   \end{algorithmic}
\end{algorithm}

\begin{figure}[t]
  \centering
  \includegraphics[width=0.98\linewidth]{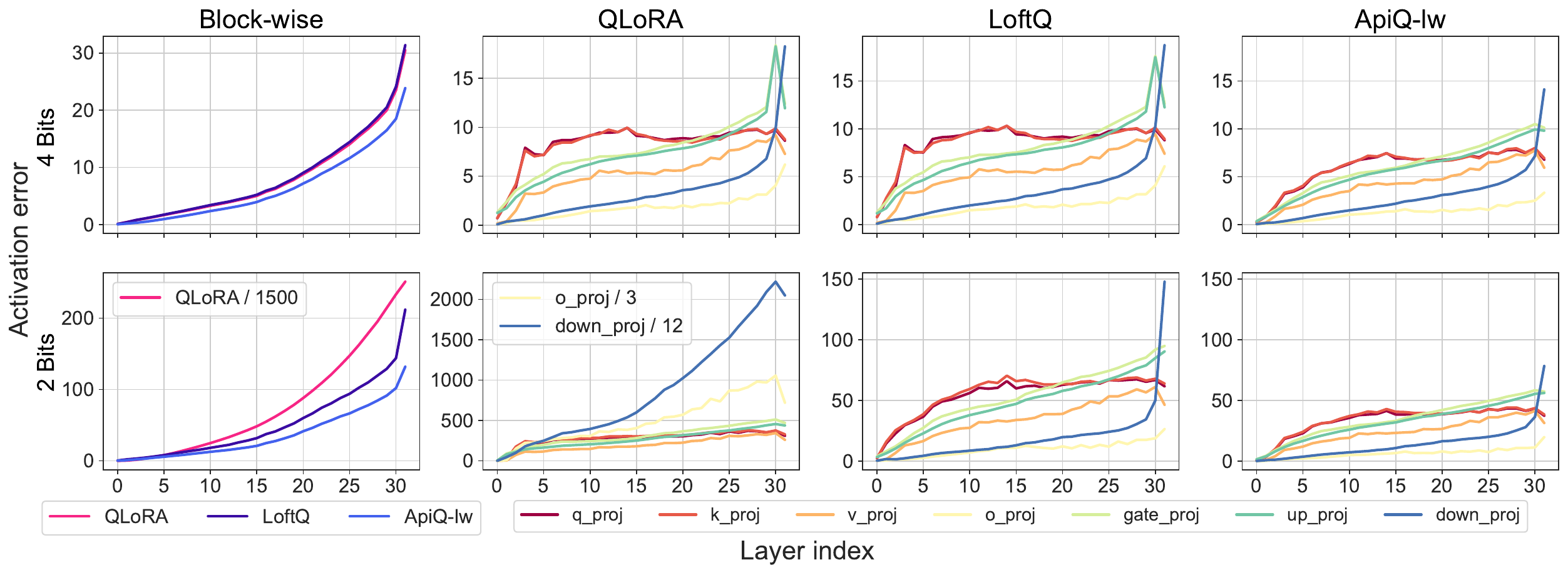}
  \caption{The average activation error $\lVert\bm{X}\bm{W} - \bm{X}^q(\bm{Q} + \bm{AB}^\top)\rVert_F$ per token for different linear layers of Llama-2-7B. \textbf{1st column}: The activation error for every transformer block. We randomly sample 128 sentences from C4 to obtain the activations. For better visualization, some lines are divided by a factor, denoted as ``/ factor''. Please pay attention to the scale of the y-axis to compare different methods. ApiQ has the smallest activation error.}
  \label{fig: act diff}
\end{figure}

\subsection{Gradient-based optimization}
To effectively solve Equation (\ref{eq: apiq}), ApiQ utilizes a gradient-based algorithm akin to conventional neural network training. The process involves optimizing the quantized weight $\bm{Q}$ along with the low-rank matrices $\bm{A}$ and $\bm{B}$ jointly.

Originally, $f$ from Equation (\ref{eq: f}) is a static function without any trainable parameters. Drawing inspiration from existing learnable quantization methods \citep{omniquant, DBLP:conf/cvpr/LiuCHXS22, DBLP:conf/iclr/EsserMBAM20, DBLP:journals/corr/abs-1805-06085}, ApiQ introduces two trainable parameters, $\gamma$ and $\beta$, for each weight matrix. These parameters control the clipping range of the quantization process:
\begin{align}
  \begin{aligned}[c]
  s &= \frac{\sigma(\gamma)\mathrm{max}(\bm{W})-\sigma(\beta)\mathrm{min}(\bm{W})}{2^b-1} \nonumber \\
  \end{aligned}
  \qquad\qquad
  \begin{aligned}[c]
  z &= -\nint{\frac{\sigma(\beta)\mathrm{min}(\bm{W})}{s}} \nonumber
  \end{aligned}
\end{align}
Here, $\sigma$ denotes a sigmoid function, constraining the clipping range to prevent excessive value expansion. We initialize $\gamma=\beta=4$ ($\sigma(4)\approx0.98$) as \citet{omniquant} to maintain the original clipping range at the beginning of quantization.

The optimization for one linear layer is outlined in Algorithm \ref{alg: apiq-lw}. During this process, only $\bm{A}$, $\bm{B}$ and $\Theta = \{\gamma, \beta\}$ are trained. Given that the quantization function $f$ incorporates a round-to-nearest operation, a straight-through estimator \citep{ste} is applied to ensure the update of $\Theta$. The ApiQ-lw algorithm is designed to be memory-efficient, optimizing each layer sequentially. This implies that any GPU capable of running the model inference can be used to quantize the model using ApiQ-lw. The outputs $\bm{Y}$ and $\bm{Y}^q$ from each layer serve as inputs to optimize the subsequent adjacent layer, ensuring efficient quantization.

ApiQ-bw employs a nearly identical optimization algorithm to ApiQ-lw, with the primary distinction being that the outputs, $\bm{y}$ and $\bm{y}^q$, are generated from a transformer block rather than a linear layer. It is worth noting that while ApiQ-bw offers improved time efficiency compared to ApiQ-lw, it necessitates marginally higher GPU memory usage due to the need to cache more activations from the layers within a transformer block.

\textbf{Preliminary experiments.} In Figure \ref{fig: act diff}, ApiQ reduces the activation error by a large margin compared to QLoRA and LoftQ, more obvious for lower-bit quantization. Interestingly, while our objective is to minimize the activation error, the weight error of ApiQ is the smallest for most layers, as shown in Figure \ref{fig: relative weight diff}. This dual effectiveness in minimizing both activation and weight errors underscores the comprehensive nature of ApiQ to quantization. Further evidence of ApiQ's effectiveness is presented in Table \ref{tab: partial lora} where ApiQ has the smallest performance gap for different trainable LoRA positions. In some cases, only training the LoRA modules in the attention position can offer the best results, similar to the original findings of LoRA \citep{lora}. It suggests that ApiQ is particularly adept at addressing and mitigating the cumulative effects of quantization error.

\section{Experiments}
\label{sec: results}
In this section, we evaluate ApiQ on the language understanding, language modeling, arithmetic reasoning and commonsense reasoning tasks by quantizing DeBERTa-v3 \cite{deberta}, RoBERTa \citep{roberta}, Llama-2 \citep{llama-2} and Mistral \citep{mistral}. Like QLoRA, LoftQ and LQ-LoRA, ApiQ consists of two steps: the quantization step and the finetuning step. During the quantization step, we initialize $\bm{Q}$, $\bm{A}$ and $\bm{B}$ in a way to preserve the starting point and mitigate the propagation of quantization error. For the finetuning step, we freeze $\bm{Q}$ in a lower bit and train $\bm{A}$ and $\bm{B}$ in half-precision (BFloat16).

\textbf{Implementation details.} In Algorithm \ref{alg: apiq-lw}, the quantization process of ApiQ requires a calibration dataset. We randomly sample 128 sentences from the training set of WikiText-2 \citep{wikitext}. Following our baselines \citep{qlora, loftq}, LoRA modules are integrated into all linear layers. By default, the group/block size for quantization is 64 for all methods. We employ AdamW \citep{adamw} as an optimizer to update $\bm{A}$, $\bm{B}$ and $\Theta$. More implementation details for the quantization and finetuning steps are detailed in Appendix \S\ref{sec: experimental details} for reproduction.

\textbf{Baselines} include full finetuning (Full FT), LoRA \citep{lora}, QLoRA \citep{qlora}, GPTQ-LoRA \citep{gptq}, LoftQ \citep{loftq}, and LQ-LoRA \citep{lq-lora}. Full FT and LoRA are considered the upper bound for finetuning. QLoRA and GPTQ-LoRA employ NF-quantization and uniform quantization, respectively, on the pretrained weights with the default LoRA initialization. These methods are memory-efficient but distort the starting point. In contrast, LoftQ and LQ-LoRA initialize the matrices $\bm{Q}$, $\bm{A}$, and $\bm{B}$ to preserve the initial weight state, thus serving as a strong baseline to ApiQ.\footnote{Notably, LoftQ and QLoRA utilize fake 3-bit quantization, applying 4-bit quantization to the first half of the shallower layers and 2-bit quantization to the remaining layers.}

\begin{table}[t]
  \centering
  \scriptsize
  \caption{The perplexity of ApiQ as a post-training quantization method without the finetuning step. The \textbf{best} and \underline{second-best} results are in bold and underlined, respectively.\label{tab: ptq}}
  \begin{tabular}{lr|rr|rr}
  \toprule
  & & \multicolumn{2}{c|}{\textbf{Llama-2-7B}} & \multicolumn{2}{c}{\textbf{Llama-2-13B}} \\
  \textbf{Method} & \textbf{Bit} & \textbf{WikiText} $\downarrow$ & \textbf{C4} $\downarrow$ & \textbf{WikiText} $\downarrow$ & \textbf{C4} $\downarrow$ \\
  \midrule
  - & 16 & 5.47 & 6.97 & 4.88 & 6.46 \\ 
  \midrule
  QLoRA & 4 & 5.65 & 7.16 & 4.98 & 6.57 \\
  LoftQ & 4 & 5.62 & 7.16 & 4.96 & \underline{6.55} \\
  ApiQ-lw & 4 & \underline{5.55} & \underline{7.08} & \underline{4.95} & \underline{6.55} \\
  ApiQ-bw & 4 & \textbf{5.53} & \textbf{7.06} & \textbf{4.93} & \textbf{6.53} \\
  \midrule
  QLoRA & 3 & 1.8e5 & 2.4e5 & 9.6e4 & 1.2e5 \\
  LoftQ & 3 & 10.72 & 12.79 & 6.89 & 8.72 \\
  ApiQ-lw & 3 & \underline{5.87} & \underline{7.58} & \underline{5.18} & \underline{6.88} \\
  ApiQ-bw & 3 & \textbf{5.77} & \textbf{7.48} & \textbf{5.12} & \textbf{6.83} \\
  \midrule
  QLoRA & 2 & 1.8e5 & 2.4e5 & 9.7e4 & 1.3e5 \\
  LoftQ & 2 & 1.0e3 & 6.7e2 & 59.94 & 72.64 \\
  ApiQ-lw & 2 & \underline{16.25} & \underline{23.93} & \underline{10.89} & \underline{15.83} \\
  ApiQ-bw & 2 & \textbf{7.59} & \textbf{10.56} & \textbf{6.44} & \textbf{8.93} \\
  \bottomrule
  \end{tabular}
  \centering
\end{table}

\begin{table}[t]
  \centering
  \scriptsize
  \caption{The comparison between ApiQ and other standard post-training quantization methods.\label{tab: compare to ptq}}
  \begin{tabular}{lrr|rr|rr}
  \toprule
  & & & \multicolumn{2}{c|}{\textbf{Llama-2-7B}} & \multicolumn{2}{c}{\textbf{Llama-2-13B}} \\
  \textbf{Method} & \textbf{Bit} & \textbf{Group size} & \textbf{WikiText} $\downarrow$ & \textbf{C4} $\downarrow$ & \textbf{WikiText} $\downarrow$ & \textbf{C4} $\downarrow$ \\
  \midrule
  - & 16 & - & 5.47 & 6.97 & 4.88 & 6.46 \\ 
  \midrule
  RTN & 4 & 128 & 5.72 & 7.24 & 4.98 & 6.58 \\
  GPTQ & 4 & 128 & 5.61 & 7.12 & 4.98 & 6.56 \\
  AWQ & 4 & 128 & 5.62 & 7.13 & 4.97 & 6.56 \\
  OmniQuant & 4 & 128 & 5.58 & 7.12 & 4.95 & 6.56 \\
  ApiQ-bw & 4 & 128 & \textbf{5.54} & \textbf{7.09} & \textbf{4.94} & \textbf{6.55} \\
  \midrule
  RTN & 3 & 128 & 6.66 & 8.40 & 5.51 & 7.18 \\
  GPTQ & 3 & 128 & 6.29 & 7.89 & 5.42 & 7.00 \\
  AWQ & 3 & 128 & 6.24 & 7.84 & 5.32 & 6.94 \\
  OmniQuant & 3 & 128 & 6.03 & 7.75 & 5.28 & 6.98 \\
  ApiQ-bw & 3 & 128 & \textbf{5.86} & \textbf{7.63} & \textbf{5.20} & \textbf{6.92}  \\
  \midrule
  RTN & 2 & 64 & 431.97 & 475.35 & 26.22 & 28.69 \\
  GPTQ & 2 & 64 & 20.85 & 19.40 & 22.44 & 12.48 \\
  AWQ & 2 & 64 & 2.1e5 & 1.6e5 & 1.2e5 & 9.5e4 \\
  OmniQuant & 2 & 64 & 9.62 & 12.72 & 7.56 & 10.05 \\
  ApiQ-bw & 2 & 64 & \textbf{7.59} & \textbf{10.56} & \textbf{6.44} & \textbf{8.93} \\
  \midrule
  RTN & 2 & 128 & 4.2e3 & 4.9e3 & 122.08 & 139.65 \\
  GPTQ & 2 & 128 & 36.77 & 33.70 & 28.14 & 20.97 \\
  AWQ & 2 & 128 & 2.2e5 & 1.7e5 & 1.2e5 & 9.4e4 \\
  OmniQuant & 2 & 128 & 11.06 & 15.02 & 8.26 & 11.05 \\
  ApiQ-bw & 2 & 128 & \textbf{8.25} & \textbf{12.04} & \textbf{6.71} & \textbf{9.13} \\
  \bottomrule
  \end{tabular}
  \centering
\end{table}

\begin{figure}[t]
  \centering
  \includegraphics[width=0.8\linewidth]{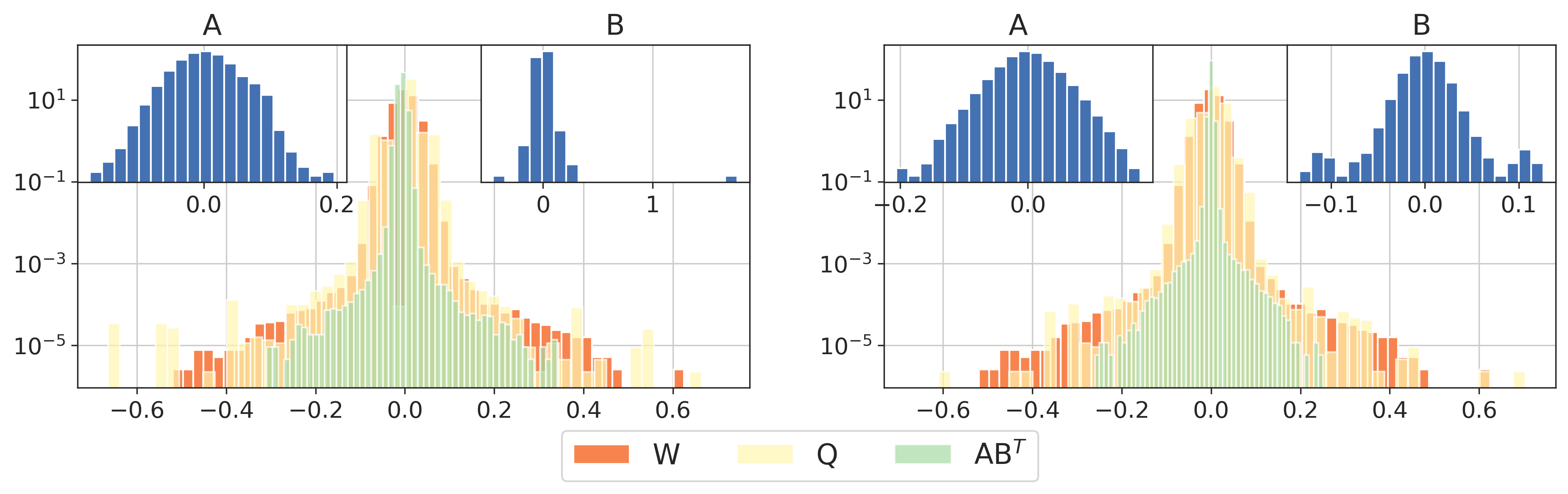}
  \caption{Histogram of $\bm{Q}$, $\bm{A}$ and $\bm{B}$ for the 2-bit quantized output projection layer in the 30$^\mathrm{th}$ block of Llama-2-7B. \textbf{Left}: LoftQ. \textbf{Right}: ApiQ-lw. Refer to Figure \ref{fig: hist 4bit}, \ref{fig: hist 3bit}, \ref{fig: hist 2bit} and \ref{fig: hist loftq 2bit} for all layers. \label{fig: hist layer}}
\end{figure}

\subsection{Quantization quality}
\label{sec: quantization quality}
In Section \S\ref{sec: method}, we have demonstrated the superior quantization quality of ApiQ by comparing the weight and activation error after quantization. Here, we further evaluate ApiQ as a post-training quantization (PTQ) method, comparing it with other PTQ methods in a language modeling task.

We begin by comparing QLoRA \citep{qlora}, LoftQ \citep{loftq} and ApiQ on the WikiText-2 test set \citep{wikitext} and the C4 validation set \citep{c4}, following the implementation details outlined in Appendix \S\ref{sec: eval of qllm}. For all methods, the quantization group size is set to 64, and the LoRA rank 
$r$ is 64. As shown in Table \ref{tab: ptq}, ApiQ-bw and ApiQ-lw consistently achieve the best and second-best perplexity across various bit levels. Notably, the performance gap between ApiQ and the baselines widens at lower bit levels.

Next, we compare ApiQ to other standard PTQ methods such as round-to-nearest quantization (RTN), GPTQ \citep{gptq}, AWQ \citep{awq}, and OmniQuant \citep{omniquant}. We exclude ApiQ-lw in this comparison as ApiQ-bw demonstrates better performance (refer to Table \ref{tab: ptq}). It is crucial to note that our objective is not to merely outperform existing PTQ methods. This is because the LoRA components in ApiQ are stored in FP16 format, inherently increasing the average bit-width per parameter, which makes direct comparisons with other PTQ methods less fair. Instead, our goal is to introduce a novel PTQ approach that mitigates quantization difficulty through the integration of LoRA components.

As illustrated in Table \ref{tab: compare to ptq}, ApiQ-bw consistently delivers the smallest perplexity, with a more significant advantage at lower bit levels. ApiQ-bw can be viewed as a combination of OmniQuant and a new initialization of LoRA, as OmniQuant employs a similar quantization algorithm as Algorithm \ref{alg: apiq-lw} without LoRA parameters. Nonetheless, ApiQ-bw outperforms OmniQuant, highlighting the effectiveness of jointly initializing the LoRA modules and quantizing the LLM weights.

A critical question arises: how does ApiQ compensate for the information loss inherent in quantization? The histograms of $\bm{Q}$, $\bm{A}$, and $\bm{B}$ in Figure \ref{fig: hist layer} provide insights into this process. Uniform quantization causes many values in $\bm{W}$ near the center to be mapped to the same value, leading to quantization error. ApiQ addresses this by centering $\bm{AB}^\top$ in this critical region. Additionally, the distribution span of ApiQ's $\bm{A}$ and $\bm{B}$ is significantly narrower compared to $\bm{W}$ and LoftQ, suggesting the potential for further quantizing $\bm{A}$ and $\bm{B}$ to reduce the overall bit-width per parameter.

\begin{table}[t]
  \centering
  \scriptsize
  \caption{The duration and peak GPU memory used for quantizing Llama-2.\label{tab: duration and mem}}
  \begin{tabular}{rlrr}
  \toprule
  \textbf{Size} & \textbf{Method} & \textbf{Duration} & \textbf{Peak GPU memory} \\
  \midrule
  & GPTQ & 0.2h & 6GB \\
  & OmniQuant & 1.1h & 12GB \\
  7B & LoftQ & 0.6h & 14GB \\
  & ApiQ-lw & 4.1h & 6GB \\
  & ApiQ-bw & 1.3h & 12GB \\
  \midrule
  & GPTQ & 0.4h & 9GB \\
  & OmniQuant & 2.2h & 16GB \\
  13B & LoftQ & 1.3h & 27GB \\
  & ApiQ-lw & 6.5h & 9GB \\
  & ApiQ-bw & 2.4h & 17GB \\
  \bottomrule
  \end{tabular}
  \centering
\end{table}

\subsection{Quantization efficiency}
\label{sec: quantization efficiency}
In this section, we compare the duration and GPU memory usage of quantization between ApiQ and other baseline methods. Detailed implementation procedures for quantization are provided in Appendix \S\ref{sec: quantization details}. It is worth noting that an LLM needs to be quantized only once and can then be saved for finetuning across various downstream tasks.

As shown in Table \ref{tab: duration and mem}, GPTQ stands out as the most efficient PTQ method, requiring the least time and GPU memory. ApiQ-lw uses a similar amount of GPU memory as GPTQ but requires more time due to its layer-by-layer sequential optimization. Similar to OmniQuant, ApiQ-bw consumes more memory than ApiQ-lw because it needs to cache more activations within a transformer block. However, ApiQ-bw is significantly more time-efficient than ApiQ-lw due to its block-by-block quantization approach. LoftQ requires the most GPU memory because of SVD. Overall, the resources required for ApiQ's quantization are reasonable and considerably lower than those needed for the finetuning step.

\subsection{Finetuning results}
\label{sec: finetuning results}
\begin{table}[t]
  \centering
  \tiny
  \caption{Results of encoder-only models on the GLUE development set. Results denoted by $^{\ast}$, $^{\dagger}$ and $^{\diamond}$ are from \citet{loftq}, \citet{lq-lora} and \citet{qlora}, respectively. When there is only one number for two metrics, it is an average over these two metric. \label{tab: glue numbers}}
  \begin{tabular}{llc|cccccccc|c}
  \toprule
  \textbf{Model} & \textbf{Method} & \textbf{Bit} & \textbf{MNLI} & \textbf{QNLI} & \textbf{QQP} & \textbf{SST-2} & \textbf{CoLA} & \textbf{RTE} & \textbf{MRPC} & \textbf{STS-B} & \textbf{Avg.} $\uparrow$ \\
  & & & m/mm & Acc & Acc/F1  & Acc & Matt & Acc & Acc/F1 & Pea/Spe & \\
  \midrule 
  & Full FT$^{\ast}$ & 16 & 90.5/90.6 & 94.0 & 92.4/89.8 & 95.3 & 69.2 & 82.0 & 89.5/93.3 & 91.6/91.1 & 88.1 \\
  \cmidrule(lr){2-12}
  & ApiQ-lw & 3 & 90.3/90.2 & 93.9 & 92.6/90.1 & 95.8 & 71.9 & 85.9 & 91.7/94.0 & 91.5/91.3 & 89.2 \\
  \cmidrule(lr){2-12}
  DeBERTa-base & QLoRA$^{\ast}$ & 2 & 79.9/79.5 & 83.7 & 88.6/84.7 & 86.9 & N.A. & 57.8 & 76.5/84.5 & 84.1/84.0 & 69.9 \\
  & LoftQ & 2 & 88.5/88.5 & 92.7 & 91.6/88.7 & 94.7 & 63.6 & 64.6 & 88.5/91.8 & 89.2/89.0 & 84.2 \\
  & ApiQ-lw & 2 & 88.4/88.7 & 92.3 & 91.7/89.0 & 94.6 & 64.2 & 67.1 & 89.5/92.4 & 90.2/89.9 & \textbf{84.8} \\
  \midrule
  & Full FT$^{\dagger}$ & 16 & 89.7 & 94.1 & 89.8 & 95.8 & 70.2 & 84.1 & 92.0 & 92.2 & 88.5 \\
  \cmidrule(lr){2-12}
  & QLoRA$^{\diamond}$ & 4 & - & - & - & - & - & - & - & - &  88.6 \\
  \cmidrule(lr){2-12}
  RoBERTa-large & QLoRA$^{\dagger}$ & 3 & 89.8 & 94.3 & 89.9 & 96.4 & 64.3 & 70.8 & 92.0 & 91.6 & 86.1 \\
  & LQ-LoRA$^{\dagger}$ & 3 & 90.3 & 94.6 & 89.7 & 96.2 & 63.5 & 80.5 & 92.2 & 91.8 & 87.3 \\
  & ApiQ-lw & 3 & 90.1/90.0 & 94.4 & 91.8/89.1 & 96.2 & 64.6 & 84.8 & 91.4/93.7 & 92.3/92.0 & \textbf{88.2} \\
  \bottomrule
  \end{tabular}
  \centering
\end{table}

\begin{table}[t]
  \centering
  \tiny
  \caption{Finetuning results of WikiText and GSM8K on Llama-2-7B, Llama-2-13B and Mistral-7B-v0.1. Results without standard deviation are from \citet{loftq}.\label{tab: wiki and gsm8k}}
  \begin{tabular}{lr|ll|ll|rr}
  \toprule
  & & \multicolumn{2}{c|}{\textbf{Llama-2-7B}} & \multicolumn{2}{c|}{\textbf{Llama-2-13B}} & \multicolumn{2}{c}{\textbf{Mistral-7B-v0.1}} \\
  \textbf{Method} & \textbf{Bit} & \textbf{WikiText} (ppl$\downarrow$) & \textbf{GSM8K} (acc$\uparrow$) & \textbf{WikiText} (ppl$\downarrow$) & \textbf{GSM8K} (acc$\uparrow$) & \textbf{WikiText} (ppl$\downarrow$) & \textbf{GSM8K} (acc$\uparrow$) \\
  \midrule
  LoRA & 16 & 5.08 & 36.9 & 5.12 & 45.3 & 5.17$_{0.00}$ & 52.2$_{1.3}$ \\
  \midrule
  QLoRA & 4 & 5.70 & 35.1 & 5.22 & 39.9 & \textbf{5.25}$_{0.00}$ & 56.5$_{1.1}$ \\
  LoftQ & 4 & \textbf{5.24} & 35.0 & \underline{5.16} & 45.0 & \textbf{5.25}$_{0.00}$ & 56.7$_{1.4}$\\
  ApiQ-lw & 4 & 5.28$_{0.00}$ & \underline{36.4}$_{0.5}$ & \textbf{4.78}$_{0.00}$ & \underline{50.4}$_{1.3}$ & 5.32$_{0.00}$ & \underline{57.2}$_{0.3}$ \\
  ApiQ-bw & 4 & \underline{5.27}$_{0.00}$ & \textbf{39.8}$_{0.1}$ & \textbf{4.78}$_{0.00}$ & \textbf{51.2}$_{0.8}$ & \underline{5.26}$_{0.00}$ & \textbf{59.2}$_{0.1}$\\
  \midrule
  QLoRA & 3 & 5.73 & 32.1 & 5.22 & 40.7 & 1540.26$_{36.6}$ & 50.5$_{0.7}$ \\
  LoftQ & 3 & 5.63 & 32.9 & 5.13 & 44.4 & 6.82$_{0.01}$ & 51.6$_{0.6}$ \\
  ApiQ-lw & 3 & \underline{5.53}$_{0.01}$ & \underline{36.0}$_{0.3}$ & \underline{4.98}$_{0.00}$ & \underline{45.4}$_{1.1}$ & \underline{5.55}$_{0.00}$ & \underline{54.8}$_{1.7}$ \\
  ApiQ-bw & 3 & \textbf{5.49}$_{0.00}$ & \textbf{39.3}$_{0.3}$ & \textbf{4.96}$_{0.00}$ & \textbf{47.6}$_{0.8}$ & \textbf{5.48}$_{0.00}$ & \textbf{56.0}$_{0.4}$\\
  \midrule
  QLoRA & 2 & N.A. & N.A. & N.A. & N.A. & 1483.56$_{12.2}$ & 2.0$_{0.3}$ \\
  LoftQ & 2 & 7.85 & 20.9 & 7.69 & 25.4 & 1849.32$_{3.78}$ & 1.7$_{0.0}$ \\
  ApiQ-lw & 2 & \underline{7.46}$_{0.00}$ & \underline{26.0}$_{0.4}$ & \underline{6.29}$_{0.00}$ & \underline{36.3}$_{0.5}$ & \underline{7.18}$_{0.00}$ & \underline{41.3}$_{0.8}$\\
  ApiQ-bw & 2 & \textbf{6.61}$_{0.00}$ & \textbf{33.5}$_{0.5}$ & \textbf{5.79}$_{0.00}$ & \textbf{41.2}$_{0.9}$ & \textbf{6.69}$_{0.00}$ & \textbf{45.0}$_{0.1}$\\
  \bottomrule
  \end{tabular}
  \centering
\end{table}
\paragraph{Natural language understanding} We finetune DeBERTa-v3-base \citep{deberta} and RoBERTa-large \citep{roberta} on the GLUE tasks \citep{glue} and show the results in Table \ref{tab: glue numbers}. ApiQ outperforms all baselines under the same level of quantization on average. With 3-bit quantization, ApiQ is even better or comparable to full finetuning.

\begin{table}[t]
  \centering
  \tiny
  \caption{Accuracy on four arithmetic reasoning tasks. All methods use the same hyper-parameters as listed in Table \ref{tab: hyper-parameter space for decoder-only}. The LoRA rank $r$ is 64 for all methods.\label{tab: arithmetic reasoning}}
  \begin{tabular}{lr|rrrr|r|rrrr|r}
  \toprule
  & & \multicolumn{5}{c|}{\textbf{Llama-2-7B}} & \multicolumn{5}{c}{\textbf{Llama-2-13B}} \\
  \textbf{Method} & \textbf{Bit} & \textbf{GSM8K} & \textbf{SVAMP} & \textbf{MAWPS} & \textbf{AQuA} & \textbf{Avg.} $\uparrow$ & \textbf{GSM8K} & \textbf{SVAMP} & \textbf{MAWPS} & \textbf{AQuA} & \textbf{Avg.} $\uparrow$ \\
  \midrule
  LoRA & 16 & 43.6$_{0.7}$ & 59.4$_{1.7}$ & 85.0$_{1.7}$ & 27.0$_{2.0}$ & 53.7$_{0.6}$ & 55.3$_{0.5}$ & 67.7$_{0.9}$ & 87.4$_{0.7}$ & 24.4$_{0.9}$ & 58.7$_{0.2}$ \\
  \midrule
  QLoRA & 4 & 42.7$_{0.4}$ & 58.7$_{0.7}$ & \textbf{87.3}$_{1.9}$ & \textbf{26.4}$_{1.6}$ & \textbf{53.7}$_{0.6}$ & 54.8$_{0.5}$ & \textbf{69.4}$_{0.3}$ & 87.0$_{0.7}$ & \textbf{26.8}$_{1.0}$ & \textbf{59.5}$_{0.3}$ \\
  GPTQ-LoRA & 4 & 43.0$_{0.9}$ & 58.4$_{0.6}$ & 86.1$_{0.7}$ & 24.3$_{0.8}$ & 52.9$_{0.3}$ & 53.2$_{0.9}$ & 67.5$_{1.2}$ & 85.3$_{0.7}$ & 25.6$_{2.6}$ & 57.9$_{1.0}$ \\
  LoftQ & 4 & 41.7$_{0.6}$ & 56.0$_{0.8}$ & 86.3$_{0.5}$ & 25.3$_{1.0}$ & 52.3$_{0.5}$ & 54.9$_{1.4}$ & 66.5$_{0.7}$ & 87.7$_{0.5}$ & 23.9$_{1.6}$ & 58.3$_{0.6}$ \\
  ApiQ-bw & 4 & \textbf{43.2}$_{0.9}$ & \textbf{59.0}$_{0.9}$ & 85.7$_{0.7}$ & 26.0$_{1.8}$ & 53.5$_{0.8}$ & \textbf{55.3}$_{0.6}$ & 67.4$_{0.5}$ & \textbf{87.8}$_{0.9}$ & 25.6$_{0.2}$ & 59.0$_{0.4}$ \\
  \midrule
  QLoRA & 3 & 1.4$_{0.2}$ & 1.4$_{0.3}$ & 0.7$_{0.5}$ & 3.4$_{1.5}$ & 1.7$_{0.5}$ & 0.8$_{0.6}$ & 2.5$_{2.2}$ & 0.3$_{0.2}$ & 6.2$_{6.8}$ & 2.4$_{2.0}$ \\
  GPTQ-LoRA & 3 & 38.9$_{0.4}$ & 55.7$_{1.2}$ & 84.9$_{0.3}$ & 23.2$_{1.6}$ & 50.7$_{0.9}$ & 50.6$_{0.0}$ & 65.2$_{1.5}$ & 88.0$_{1.0}$ & 22.6$_{1.3}$ & 56.6$_{0.8}$ \\
  LoftQ & 3 & 39.9$_{0.4}$ & \textbf{56.3}$_{2.2}$ & 86.3$_{0.8}$ & \textbf{26.4}$_{1.4}$ & 52.2$_{0.7}$ & \textbf{53.9}$_{1.2}$ & 66.1$_{0.2}$ & 87.0$_{0.9}$ & 23.6$_{0.7}$ & 57.7$_{0.3}$ \\
  ApiQ-bw & 3 & \textbf{41.4}$_{1.5}$ & 55.9$_{0.3}$ & \textbf{87.0}$_{1.4}$ & 25.2$_{0.9}$ & \textbf{52.4}$_{0.6}$ & 51.5$_{0.8}$ & \textbf{67.4}$_{0.3}$ & \textbf{88.5}$_{1.2}$ & \textbf{25.6}$_{1.3}$ & \textbf{58.3}$_{0.3}$ \\
  \midrule
  QLoRA & 2 & 0.9$_{0.4}$ & 1.5$_{1.1}$ & 0.8$_{0.7}$ & 5.1$_{4.9}$ & 2.1$_{1.7}$ & 0.5$_{0.4}$ & 0.7$_{0.9}$ & 0.1$_{0.2}$ & 0.9$_{1.3}$ & 0.6$_{0.4}$ \\
  GPTQ-LoRA & 2 & 21.7$_{0.6}$ & 39.0$_{1.3}$ & 76.6$_{0.8}$ & 22.1$_{1.8}$ & 39.9$_{0.5}$ & 31.9$_{0.0}$ & 49.6$_{1.0}$ & 82.5$_{0.4}$ & \textbf{23.6}$_{0.9}$ & 46.9$_{0.5}$ \\
  LoftQ & 2 & 29.5$_{0.8}$ & 45.8$_{0.7}$ & \textbf{83.6}$_{0.6}$ & 23.2$_{2.0}$ & 45.6$_{0.7}$ & 37.0$_{0.6}$ & 55.9$_{0.8}$ & \textbf{87.7}$_{1.3}$ & 21.7$_{1.1}$ & 50.6$_{0.2}$ \\
  ApiQ-bw & 2 & \textbf{31.2}$_{0.5}$ & \textbf{51.0}$_{1.1}$ & 82.9$_{1.6}$ & \textbf{23.9}$_{1.0}$ & \textbf{47.3}$_{0.5}$ & \textbf{43.1}$_{0.8}$ & \textbf{59.2}$_{1.2}$ & 85.1$_{1.1}$ & 23.4$_{1.4}$ & \textbf{52.7}$_{0.5}$ \\
  \bottomrule
  \end{tabular}
  \centering
\end{table}

\begin{table}[t]
  \centering
  \tiny
  \caption{Accuracy on eight commonsense reasoning tasks. All methods use the same hyper-parameters as listed in Table \ref{tab: hyper-parameter space for decoder-only}. The LoRA rank $r$ is 64 for all methods.\label{tab: commonsense reasoning results}}
  \begin{tabular}{llr|rrrrrrrr|r}
  \toprule
  \textbf{Model} & \textbf{Method} & \textbf{Bit} & \textbf{BoolQ} & \textbf{PIQA} & \textbf{SIQA} & \textbf{HellaS.} & \textbf{WinoG.} & \textbf{ARC-e} & \textbf{ARC-c} & \textbf{OBQA} & \textbf{Avg.} $\uparrow$ \\
  \midrule
  & LoRA & 16 & 73.6$_{0.3}$ & 86.5$_{0.1}$ & 81.8$_{0.1}$ & 95.2$_{0.1}$ & 86.9$_{0.2}$ & 89.4$_{0.4}$ & 76.7$_{0.8}$ & 86.7$_{0.8}$ & 84.6$_{0.2}$ \\
  \cmidrule(lr){2-12}
  & QLoRA & 4 & \textbf{73.9}$_{0.4}$ & 84.4$_{0.8}$ & 79.7$_{0.2}$ & 93.3$_{0.2}$ & 84.6$_{0.6}$ & 86.1$_{0.4}$ & 73.0$_{0.4}$ & 85.1$_{0.5}$ & 82.5$_{0.2}$ \\
  & GPTQ-LoRA & 4 & 73.4$_{0.4}$ & 83.6$_{0.5}$ & 79.3$_{0.3}$ & 93.3$_{0.1}$ & 84.5$_{0.8}$ & 86.5$_{0.3}$ & 72.8$_{1.3}$ & 83.3$_{0.2}$ & 82.1$_{0.2}$\\
  & LoftQ & 4 & 73.7$_{0.4}$ & 86.0$_{0.6}$ & 81.1$_{0.2}$ & 94.6$_{0.3}$ & 86.3$_{0.1}$ & 88.1$_{0.5}$ & 75.5$_{1.0}$ & \textbf{86.2}$_{0.6}$ & 83.9$_{0.2}$\\
  & ApiQ-bw & 4 & 73.5$_{0.2}$ & \textbf{87.0}$_{0.4}$ & \textbf{82.0}$_{0.1}$ & \textbf{95.2}$_{0.1}$ & \textbf{86.9}$_{0.2}$ & \textbf{89.5}$_{0.4}$ & \textbf{77.0}$_{0.8}$ & \textbf{86.2}$_{0.4}$ & \textbf{84.7}$_{0.2}$\\
  \cmidrule(lr){2-12}
  Llama-2-7B & GPTQ-LoRA & 3 & 71.8$_{0.2}$ & 82.7$_{0.3}$ & 79.3$_{0.6}$ & 92.1$_{0.1}$ & 82.8$_{0.3}$ & 84.2$_{0.6}$ & 70.6$_{0.8}$ & 83.4$_{1.1}$ & 80.8$_{0.1}$\\
  & LoftQ & 3 & \textbf{74.0}$_{0.0}$ & \textbf{85.6}$_{0.4}$ & 81.0$_{0.7}$ & 94.3$_{0.1}$ & 85.6$_{0.1}$ & \textbf{88.1}$_{0.6}$ & \textbf{75.4}$_{0.8}$ & 85.5$_{0.7}$ & 83.7$_{0.3}$\\
  & ApiQ-bw & 3 & 73.3$_{0.3}$ & \textbf{85.6}$_{0.1}$ & \textbf{81.8}$_{0.5}$ & \textbf{94.6}$_{0.0}$ & \textbf{86.9}$_{0.5}$ & 87.9$_{0.3}$ & 73.7$_{0.3}$ & \textbf{86.4}$_{1.3}$ & \textbf{83.8}$_{0.1}$ \\
  \cmidrule(lr){2-12}
  & GPTQ-LoRA & 2 & 62.2$_{0.0}$ & 49.5$_{0.2}$ & 33.3$_{0.6}$ & 25.1$_{0.1}$ & 49.4$_{0.4}$ & 25.0$_{0.2}$ & 22.6$_{0.0}$ & 27.6$_{0.0}$ & 36.8$_{0.0}$\\
  & LoftQ & 2 & 62.4$_{0.0}$ & 70.5$_{2.9}$ & 73.4$_{0.5}$ & 78.8$_{3.6}$ & 71.0$_{3.7}$ & 66.5$_{4.5}$ & 50.8$_{4.3}$ & 62.3$_{7.5}$ & 67.0$_{3.3}$ \\
  & ApiQ-bw & 2 & \textbf{68.4}$_{0.7}$ & \textbf{80.7}$_{0.3}$ & \textbf{79.6}$_{0.5}$ & \textbf{91.4}$_{0.1}$ & \textbf{82.4}$_{0.5}$ & \textbf{82.7}$_{0.8}$ & \textbf{68.3}$_{0.6}$ & \textbf{80.5}$_{0.6}$ & \textbf{79.3}$_{0.2}$\\
  \midrule
  & LoRA & 16 & 76.3$_{0.2}$ & 88.5$_{0.0}$ & 83.4$_{0.3}$ & 96.5$_{0.2}$ & 89.6$_{0.4}$ & 92.8$_{0.4}$ & 81.7$_{0.4}$ & 89.6$_{0.4}$ & 87.3$_{0.1}$\\
  \cmidrule(lr){2-12}
  & QLoRA & 4 & 74.9$_{0.5}$ & 86.6$_{0.5}$ & 81.5$_{0.5}$ & 94.9$_{0.1}$ & 86.9$_{0.2}$ & 89.1$_{0.7}$ & 77.1$_{0.4}$ & 87.2$_{0.7}$ & 84.8$_{0.3}$\\
  & GPTQ-LoRA & 4 & 74.5$_{0.6}$ & 86.1$_{1.0}$ & 81.8$_{0.2}$ & 94.7$_{0.3}$ & 86.8$_{0.1}$ & 89.0$_{0.1}$ & 77.1$_{0.9}$ & 84.5$_{1.2}$ & 84.3$_{0.0}$ \\
  & LoftQ & 4 & 76.0$_{0.3}$ & 87.9$_{0.2}$ & 82.8$_{0.6}$ & 95.8$_{0.1}$ & 88.9$_{0.6}$ & 91.2$_{0.3}$ & 80.8$_{0.7}$ & 88.8$_{1.3}$ & 86.5$_{0.2}$ \\
  & ApiQ-bw & 4 & \textbf{76.2}$_{0.3}$ & \textbf{88.5}$_{0.3}$ & \textbf{83.5}$_{0.1}$ & \textbf{96.6}$_{1.4}$ & \textbf{90.0}$_{0.4}$ & \textbf{92.1}$_{0.1}$ & \textbf{81.2}$_{0.3}$ & \textbf{89.9}$_{0.5}$ & \textbf{87.3}$_{0.1}$ \\
  \cmidrule(lr){2-12}
  Llama-2-13B & GPTQ-LoRA & 3 & 73.5$_{0.5}$ & 85.2$_{0.3}$ & 81.1$_{0.5}$ & 94.1$_{0.1}$ & 85.7$_{0.3}$ & 87.9$_{0.4}$ & 75.5$_{0.7}$ & 85.3$_{0.9}$ & 83.5$_{0.0}$\\
  & LoftQ & 3 & 75.2$_{0.3}$ & 87.8$_{0.6}$ & \textbf{82.8}$_{0.2}$ & \textbf{96.3}$_{0.1}$ & \textbf{89.5}$_{0.4}$ & \textbf{91.1}$_{0.1}$ & \textbf{81.4}$_{0.5}$ & 88.0$_{0.5}$ & 86.5$_{0.2}$\\
  & ApiQ-bw & 3 & \textbf{76.0}$_{0.4}$ & \textbf{88.0}$_{0.5}$ & 82.3$_{0.1}$ & 95.8$_{0.0}$ & 89.1$_{0.1}$ & \textbf{91.1}$_{0.2}$ & 81.1$_{0.5}$ & \textbf{89.5}$_{0.4}$ & \textbf{86.6}$_{0.0}$ \\
  \cmidrule(lr){2-12}
  & GPTQ-LoRA & 2 & 62.2$_{0.0}$ & 50.1$_{0.9}$ & 34.0$_{0.6}$ & 25.1$_{0.1}$ & 49.6$_{0.4}$ & 25.0$_{0.0}$ & 22.7$_{0.0}$ & 27.6$_{0.0}$ & 37.1$_{0.3}$\\
  & LoftQ & 2 & 65.9$_{0.1}$ & 76.4$_{0.3}$ & 78.0$_{0.5}$ & 84.4$_{0.7}$ & 76.1$_{0.4}$ & 75.1$_{0.1}$ & 60.1$_{0.4}$ & 72.7$_{1.4}$ & 73.6$_{0.2}$\\
  & ApiQ-bw & 2 & \textbf{73.1}$_{0.4}$ & \textbf{85.2}$_{0.5}$ & \textbf{82.3}$_{0.5}$ & \textbf{94.4}$_{0.1}$ & \textbf{86.2}$_{0.3}$ & \textbf{88.2}$_{0.3}$ & \textbf{74.9}$_{0.4}$ & \textbf{85.9}$_{1.4}$ & \textbf{83.8}$_{0.3}$\\
  \bottomrule
  \end{tabular}
  \centering
\end{table}

\paragraph{Language modeling} We finetune Llama-2-7B, Llama-2-13B \citep{llama-2} and Mistral-7B-v0.1 \cite{mistral} on the WikiText-2 training set \citep{wikitext} and report their perplexity on the validation set, as shown in Table \ref{tab: wiki and gsm8k}. Among the tested methods, ApiQ-bw consistently achieved the best performance, followed closely by ApiQ-lw across most bit levels. The performance difference becomes more pronounced at lower bit levels. The ApiQ's results on Llama-2-13B are even better than LoRA (Float16) for the 3-bit and 4-bit levels.

\paragraph{Arithmetic reasoning (single-task)} We finetune Llama-2 and Mistral on the training set of GSM8K \citep{gsm8k} and report the accuracy on the test set in Table \ref{tab: wiki and gsm8k}. Similar to the results of WikiText-2, ApiQ-bw and ApiQ-lw achieve the highest and second-highest accuracy for all bit levels, respectively, with both ApiQs being comparable to or even better than LoRA for the 3- and 4-bit.

\paragraph{ApiQ-lw or ApiQ-bw} ApiQ-lw is more memory-efficient for quantization than ApiQ-bw (\S\ref{sec: quantization efficiency}). However, ApiQ-lw requires more time for quantization due to the layer-by-layer manner. Based on the quantization quality (\S\ref{sec: quantization quality}), quantization efficiency (\S\ref{sec: quantization efficiency}) and the previously discussed finetuning results, we recommend using ApiQ-bw, with ApiQ-lw being ignored for the following experiments.

\paragraph{Arithmetic reasoning} The setting here contrasts with the previous experiments where each task involves finetuning a separate QLLM. Instead, we adopt a unified strategy by finetuning a single QLLM across all tasks as delineated in \citet{llmadapter}. We finetune Llama-2 on Math10K \citep{llmadapter}, and evaluate the finetuned QLLM on the test sets of AQuA \citep{aqua}, GSM8K, MAWPS \citep{mawps} and SVAMP \citep{svamp}. Such a setting is more practical as LLM is frequently used as a general model for various tasks.

As shown in Table \ref{tab: arithmetic reasoning}, ApiQ-bw consistently outperforms all quantization baselines for various bit levels, except for the 4-bit level where ApiQ is slightly worse than QLoRA, 53.5 vs. 53.7 for Llama-2-7B and 59.0 vs. 59.5 for Llama-2-13B. However, QLoRA's 3- and 2-bit results are extremely worse, < 3\% accuracy.

\paragraph{Commonsense reasoning} In assessing the capacity of QLLM for commonsense reasoning, we focus on eight representative tasks: BoolQ \citep{boolq}, PIQA \citep{piqa}, SIQA \citep{siqa}, HellaSwag \citep{hellaswag}, WinoGrande \citep{winogrande}, ARC-e, ARC-c \citep{arc}, and OBQA \citep{obqa}. Similar to the multiple arithmetic reasoning tasks, 
\begin{wrapfigure}{r}{0.33\textwidth}
    \includegraphics[width=\linewidth]{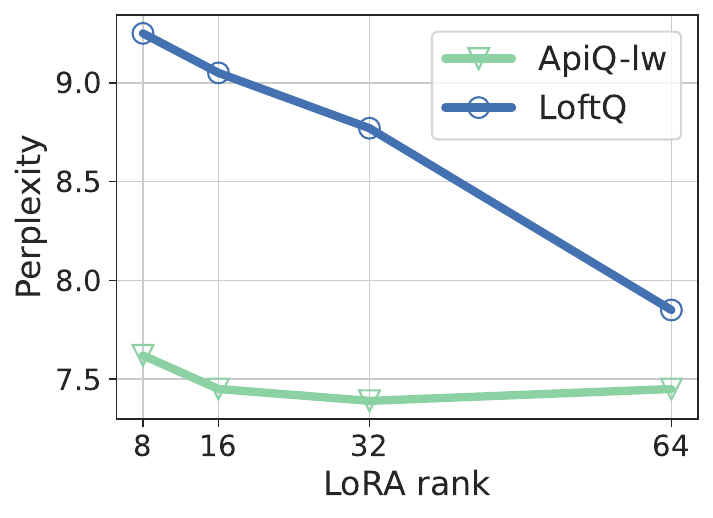} 
    \caption[]{Perplexity on WikiText-2 with 2-bit quantized Llama-2-7B for different LoRA ranks.\label{fig: lora rank}}
\end{wrapfigure}
we follow the setting of \citet{llmadapter}, finetune a single QLLM on the combined training sets from these tasks, and report the accuracy of the test sets.

As shown in Table \ref{tab: commonsense reasoning results}, ApiQ-bw consistently achieves the best average accuracy. For the 4-bit quantization, ApiQ-bw is the only method comparable to LoRA in Float16. For the 2-bit quantization, ApiQ-bw outperforms both GPTQ-LoRA and LoftQ by a large margin with an average accuracy improvement > 10\%.

\section{Discussion}
\label{discussion}
\begin{table}[t]
  \centering
  \scriptsize
  \caption{ApiQ-bw with DoRA on the commonsense reasoning tasks with 2-bit quantized Llama-2-7B. QDoRA here means that we use QLoRA to quantize the LLM, initialize the LoRA module with default $\bm{B}=\bm{0}$, and train QLLM in the DoRA way, i.e. training the direction and magnitude separately. LoftQ can't be directly applied to DoRA, because DoRA has both addition and multiplication relation between the PEFT parameters and $\bm{W}$, and SVD can't be applied. \label{tab: dora commonsense}}
  \begin{tabular}{l|rrrrrrrr|r}
  \toprule
  \textbf{Method} & \textbf{BoolQ} & \textbf{PIQA} & \textbf{SIQA} & \textbf{HellaS.} & \textbf{WinoG.} & \textbf{ARC-e} & \textbf{ARC-c} & \textbf{OBQA} & \textbf{Avg.} $\uparrow$ \\
  \midrule
  QDoRA & 62.2 & 49.7 & 33.2 & 24.4 & 48.8 & 24.4 & 22.8 & 27.2 & 36.6 \\
  ApiQ-bw with DoRA & 68.7 & 78.8 & 76.9 & 85.5 & 79.8 & 78.5 & 62.8 & 78.4 & 76.2 \\
  \bottomrule
  \end{tabular}
  \centering
\end{table}

\begin{table}[t]
  \centering
  \scriptsize
  \caption{ApiQ-bw with DoRA on the arithmetic reasoning tasks with 2-bit quantized Llama-2-7B. \label{tab: dora arithmetic}}
  \begin{tabular}{l|rrrr|r}
  \toprule
  \textbf{Method} & \textbf{GSM8K} & \textbf{SVAMP} & \textbf{MAWPS} & \textbf{AQuA} & \textbf{Avg.} $\uparrow$  \\
  \midrule
  QDoRA & 0.68 & 0.5 & 1.7 & 2.8 & 1.4 \\
  ApiQ-bw for DoRA & 32.0 & 49.9 & 80.1 & 23.4 & 46.4 \\
  \bottomrule
  \end{tabular}
  \centering
\end{table}

\paragraph{ApiQ-bw for other PEFT} As discussed in Section \S\ref{sec: apiq-bw}, ApiQ-bw can easily be applied to other PEFT methods, because its block-by-block quantization manner is very friendly to them. Here we apply ApiQ-bw to a recent variant of LoRA, i.e. DoRA \citep{dora}, and show the finetuning results in Table \ref{tab: dora commonsense} and \ref{tab: dora arithmetic}. ApiQ-bw with DoRA outperforms QDoRA by a large margin, on average 76.2 vs. 36.6 for commonsense reasoning and 46.4 v.s. 1.4 for arithmetic reasoning.

\paragraph{Performance vs. LoRA rank} In Figure \ref{fig: lora rank}, we show the influence of LoRA rank for different methods. ApiQ is not sensitive to the LoRA rank, implying that ApiQ can be a more parameter-efficient finetuning method.

\paragraph{Why ApiQ works so well} Here we discuss the reasons for the effectiveness of ApiQ. The first reason is the smaller activation error of ApiQ. Compared to LoftQ and QLoRA, ApiQ's activation error is way more smaller. As shown in Table \ref{tab: ptq}, ApiQ has a much smaller perplexity. 
Maintaining a small activation error means that the learned knowledge from the full-precision LLM is preserved, thus facilitating the transfer learning for downstream tasks.

However, maintaining a smaller activation error is not the only reason for better finetuning results. Compared to LoftQ in Table \ref{tab: ptq}, GPTQ has a smaller perplexity in Table \ref{tab: compare to ptq}. For example, GPTQ's perplexity is 20.85 for WikiText-2 on the 2-bit quantized Llama-2-7B, while LoftQ's perplexity is larger than 1000. Nevertheless, GPTQ-LoRA's \footnote{GPTQ-LoRA is LLM quantized by GPTQ and the LoRA's $\bm{B}=\bm{0}$.} finetuning results are worse than LoftQ's, e.g. 39.9 vs. 45.6 for arithmetic reasoning and 36.8 vs. 67.0 for commonsense reasoning. 

We hypothesize that the second reason is the better initialization of $\bm{A}$ and $\bm{B}$. The default initialization of $\bm{B}=\bm{0}$ is not friendly to training because of the constant value. $\bm{A}$ and $\bm{B}$ in ApiQ and LoftQ are initialized similarly to a Gaussian distribution (Figure \ref{fig: hist layer}), which has been shown better for training \citep{DBLP:conf/iccv/HeZRS15, DBLP:journals/jmlr/GlorotB10}.

\section{Related work}
\label{sec: related work}
\textbf{Large language models}, trained on web-scale data for general tasks like masked word prediction \citep{m3, bert} or next-word prediction \citep{gpt} in sentences, are crucial for transferring knowledge to various downstream tasks. These models have consistently achieved state-of-the-art results in a wide range of applications. Notably, scaling up LLMs has been observed to reliably improve performance in these downstream tasks. As a result, the size of LLMs has been steadily increasing, now reaching the remarkable scale of $>50$ billion parameters \citep{mixtral, llama-2, mistral, opt}. In addition, instruction-finetuned LLMs \citep{mistral, llama-2} reveal exceptional capabilities, such as enabling zero-shot or in-context learning \citep{radford2019language, gpt}. 

Despite these advancements, transfer learning remains the predominant strategy for effectively applying these models to new task environments \citep{ia3, gpt}. This approach, however, imposes unprecedented demands on computational resources, highlighting the need for efficient adaptation strategies. ApiQ reduces the GPU memory requirement for finetuning by loading the LLM's weights in a reduced bit format and reducing the number of trainable parameters. In addition, compared to QLoRA \citep{qlora} and its variants \citep{loftq, lq-lora}, ApiQ demonstrates a very good manner for challenging lower-bit quantization, like 2 or 3 bits, which further reduces the GPU memory. 

\textbf{Post-training quantization} (PTQ) converts high-precision LLM's weight values into discrete values for less memory usage. With the increasing size of LLMs, various PTQ methods \citep{smoothquant, awq, omniquant, spqr, gptq} have been proposed to retain the full-precision LLM's performance while using less memory during inference. Notably, PTQ aims to maintain the performance of LLMs instead of adapting LLMs to new tasks. In addition, the performance of PTQ for lower bit-width degrades significantly. Even though ApiQ is a well-behaved PTQ method, its main purpose is to adapt LLMs to new tasks or retain full-precision LLM's performance for lower-bit quantization. 

\textbf{Quantization-aware training} (QAT) is a technique where the model is trained to take into account the effects of quantization, typically reducing the precision of the model's parameters, to ensure minimal loss in performance when the model is deployed in a resource-constrained environment \citep{DBLP:conf/iclr/TailorFL21, DBLP:conf/icml/NagelFBB22}. Although QAT can be employed to adapt LLM to downstream tasks, it is memory-intensive because it involves quantization and full finetuning at the same time. In addition, some techniques, like straight-through estimator \citep{DBLP:conf/cvpr/LiuCHXS22}, are required during full finetuning to calculate the gradients, being unstable for the training of LLM. In contrast, ApiQ separates the quantization and finetuning steps, making the finetuning stable, efficient and effective. 

\textbf{Knowledge distillation} is a technique for improving the performance of small models (student model) by transferring knowledge from large and complex models (teacher model) \citep{DBLP:conf/emnlp/LiaoGN20, DBLP:journals/corr/HintonVD15, DBLP:conf/interspeech/LiZHG14, bucilu2006model}. ApiQ can be considered as a layer-wise knowledge distillation \citep{DBLP:conf/icml/LiangZZHCZ23}, where the teacher model is the full-precision LLM and the student model is the QLLM. However, ApiQ requires much fewer samples, i.e. 128 calibration sentences, to transfer the knowledge. In addition, ApiQ conducts the distillation in a layer/block once, which doesn't consume much GPU memory.

\section{Conclusion}
In this work, we propose ApiQ, a novel framework that aims to reduce the activation error during quantization by jointly quantizing the LLM's weights and initializing the LoRA's components. Extensive experiments, on five tasks with various encoder-only and decoder-only models, demonstrate ApiQ's effectiveness in adapting LLM. It works extremely well with lower-bit quantization and larger models than the strong baselines. Further experiments also demonstrate ApiQ's capability as a pure post-training quantization method.

\section*{Limitations}
\label{sec: limitation}
Although ApiQ demonstrates impressive finetuning results, some limitations are inherited from its implementation. Compared to LoftQ \citep{loftq}, ApiQ requires a calibration dataset to determine the clipping range of $\bm{W}$ and to initialize $\bm{A}$ and $\bm{B}$. This implementation has one obvious drawback: It requires more time for quantization, as shown in Table \ref{tab: duration and mem}. Since we only need to quantize the LLM once for finetuning various tasks, and the duration and GPU memory used for quantization are reasonable, we deem this limitation acceptable.

Secondly, we only evaluate ApiQ on a limited number of tasks with a total number of 5 models due to time and resource limitations. We couldn't guarantee its effectiveness on the other tasks and LLMs, and are still working on including more tasks and models, trying to show its generalization. For the post-training quantization results (\S\ref{sec: quantization quality}), we didn't make sure ApiQ shares the same bit per parameter as our baselines, which makes the direct comparison unfair. Since the main focus of this research is about finetuning, we add these results mainly to raise attention to this new method for PTQ. In the future, we aim to apply ApiQ to quantize both weight and activation for faster inference.

\section*{Acknowledgements}
We thank all reviewers of ICML2024 for their helpful feedback about block-wise ApiQ and more benchmarks. We also thank eBay Inc. for the computation support. This research was funded in part by the Netherlands Organization for Scientific Research (NWO) under project number VI.C.192.080.

\newpage
{\footnotesize
\bibliography{refs}}

\newpage
\appendix
\counterwithin{figure}{section}
\counterwithin{table}{section}

\section{Experimental Details}
\label{sec: experimental details}
Like QLoRA \citep{qlora}, LoftQ \citep{loftq} and LQ-LoRA \citep{lq-lora}, ApiQ consists of two steps: the quantization step and the finetuning step. During the quantization step, we initialize $\bm{Q}$, $\bm{A}$ and $\bm{B}$ in a way to preserve the starting point and mitigate the propagation of quantization error. For the finetuning step, we freeze $\bm{Q}$ in a lower bit and train $\bm{A}$ and $\bm{B}$ in half-precision (BFloat16). In this section, we describe the implementation details of these two steps for different tasks and LLMs. We run all experiments on NVIDIA A100-80GB or A6000-48GB with the training framework, Transformers \citep{transformers_huggingface}.

\subsection{Quantization details for different LLMs}
\label{sec: quantization details}
For all LLMs, 128 calibration sentences for the quantization step are randomly selected from the WikiText-2 training set \citep{wikitext}. The hyper-parameters for the quantization step are detailed in Table \ref{tab: quantization hyperparameter}. By default, we incorporate the LoRA module into every linear layer.

\begin{table}[ht]
  \centering
  \scriptsize
  \caption{Hyper-parameter search space of the quantization step on different LLMs. Since ApiQ-bw is more time-efficient than ApiQ-lw, we conducted a more thorough search for ApiQ-bw. The best setting for ApiQ-bw is listed in Table \ref{tab: best quantization hyperparameter}.\label{tab: quantization hyperparameter}}
  \begin{tabular}{lccc}
  \toprule
  \textbf{Hyper-parameter} & \textbf{DeBERTa} \& \textbf{RoBERTa} & \textbf{Llama-2} \& \textbf{Mistral} & \textbf{Llama-2} \& \textbf{Mistral} \\
  \midrule
  ApiQ choice & ApiQ-lw & ApiQ-lw & ApiQ-bw \\
  \midrule
  Optimizer & AdamW & AdamW & AdamW \\
  Weight decay for $\Theta$ & 0.1 & 0.1 & \{0, 0.001, 0.1\} \\
  Static LR for $\Theta$ & 0.005 & 0.005 &  \{0.001, 0.005, 0.01, 0.05\} \\
  Weight decay for $\bm{A}$ and $\bm{B}$ & 0.1 & 0.1 & \{0, 0.001, 0.1\} \\
  Static LR for $\bm{A}$ and $\bm{B}$ & 0.001 & 0.001 & \{0.0001, 0.0005, 0.001, 0.005\} \\
  Sequence length for calibration & 128 & 1024 & 2048 \\
  Number of calibration samples & 128 & 128 & 128 \\
  Epochs & 20 & 20 & 20\\
  Batch size & 32 & 8 & 1 \\
  \midrule
  Group/block size for quantization & 64 & 64 & 64 \\
  LoRA rank $r$ & 32 \& 64 & 64 & 64 \\
  \bottomrule
  \end{tabular}
  \centering
\end{table}

\begin{table*}[ht]
  \centering
  \scriptsize
  \caption{Best hyper-parameter setting for different LLMs with ApiQ-bw. If one wants to apply ApiQ-bw to other LLMs, the settings from Llama-2-7B are universally well-performed and should be the first choice. \label{tab: best quantization hyperparameter}}
  \begin{tabular}{l|lll|lll|lll}
  \toprule
  & \multicolumn{3}{c|}{\textbf{Llama-2-7B}} & \multicolumn{3}{c|}{\textbf{Llama-2-13B}} & \multicolumn{3}{c}{\textbf{Mistral-7B-v0.1}} \\
  \textbf{Hyper-parameter} & \textbf{4 Bits} & \textbf{3 Bits} & \textbf{2 Bits} & \textbf{4 Bits} & \textbf{3 Bits} & \textbf{2 Bits} & \textbf{4 Bits} & \textbf{3 Bits} & \textbf{2 Bits} \\
  \midrule
  Weight decay for $\Theta$ & 0.001  & 0.1 & 0.1 & 0.001 & 0.1 & 0.1 & 0.001 & 0.1 & 0.1 \\
  Static LR for $\Theta$ & 0.05 & 0.001 & 0.005 & 0.01 & 0.001 & 0.005 & 0.01 & 0.001 & 0.005 \\
  Weight decay for $\bm{A}$ and $\bm{B}$ & 0.1 & 0 & 0.1 & 0 & 0 & 0.1 & 0 & 0 & 0.1 \\
  Static LR for $\bm{A}$ and $\bm{B}$ & 0.0001 & 0.0005 & 0.0005 & 0.0001 & 0.0005 & 0.0005 & 0.0001 & 0.0001 & 0.0005 \\
  \bottomrule
  \end{tabular}
  \centering
\end{table*}

\textbf{DeBERTa and RoBERTa with ApiQ-lw.} We apply ApiQ-lw to DeBERTa-v3-base \citep{deberta} and RoBERTa-large \citep{roberta}, as these models are relatively small and efficiency concerns in quantization are minimal. Specifically, the duration for ApiQ-lw is 12 minutes for DeBERTa-v3-base and 1 hour for RoBERTa-large. The LoRA rank $r$ is set to 32 for DeBERTa-v3-base, following \citet{loftq}, and 64 for RoBERTa-large, as per \citet{lq-lora}. Given the relative simplicity of the GLUE tasks \citep{glue}, we only employ 2-bit and 3-bit quantization. Unlike \citet{loftq}, we do not quantize the embedding layer and instead reproduce their experiments.

\textbf{Llama-2 and Mistral with ApiQ-lw.} We apply ApiQ-lw to Llama-2-7B, Llama-2-13B \citep{llama-2} and Mistral-7B-v0.1 \citep{mistral}, with settings detailed in Table \ref{tab: quantization hyperparameter}.

\textbf{Llama-2 and Mistral with ApiQ-bw.} We also apply ApiQ-bw to Llama-2-7B, Llama-2-13B and Mistral-7B-v0.1, with the settings outlined in Table \ref{tab: quantization hyperparameter}. Compared to ApiQ-lw, we conduct an extensive search for the optimal learning rate and weight decay due to the time efficiency of ApiQ-bw. For instance, Llama-2-7B with ApiQ-lw requires 4 hours, whereas Llama-2-7B with ApiQ-bw requires only 1 hour.

To determine the best hyper-parameters, we evaluate the QLLM on the WikiText-2 test set \citep{wikitext} and the C4 validation set \citep{c4}, similar to the evaluation of post-training QLLM (see \S\ref{sec: eval of qllm}). The optimal hyper-parameter settings, determined by the lowest average perplexity across these two datasets, are listed for different LLMs in Table \ref{tab: best quantization hyperparameter}.

\subsection{Evaluation of QLLM}
\label{sec: eval of qllm}
To assess the effectiveness of quantization, we adhere to the evaluation approach used in post-training quantization methods \citep{gptvq, omniquant, smoothquant, gptq, llmint8}. For the WikiText-2 test set \citep{wikitext}, we apply the QLLM to all sentences and calculate the average perplexity. For the validation set of C4 \citep{c4}, we use the ``en/c4-validation.00000-of-00008.json.gz'' split, concatenate all sentences, randomly cut off 256 sentences with a sequence length of 2048, and compute the average perplexity using the QLLM on these samples.

\subsection{Natural language understanding}
\label{sec: nlu details}
To study the language understanding ability of LLMs, we finetune quantized DeBERTa-v3-base \cite{deberta} and RoBERTa-large \cite{roberta} on the GLUE benchmark \cite{glue}.

\textbf{Finetuning details.} The hyper-parameters for finetuning are outlined in Table \ref{tab: glue hyperparameter}. We save the checkpoint every epoch, evaluate it on the development set, and report the best result. After deciding the best learning rate, three random runs are conducted and the median is reported in Table \ref{tab: glue numbers}.

\begin{table}[ht]
  \centering
  \footnotesize
  \caption{Hyper-parameter search space for the finetuning on GLUE. For tasks with a number of training samples > 10K, we set the number of epochs as 10.\label{tab: glue hyperparameter}}
  \begin{tabular}{lcc}
  \toprule
  \textbf{Hyper-parameter} & \textbf{RTE, MRPC, STS-B, CoLA} & \textbf{SST-2, QNLI, QQP, MNLI} \\
  \midrule
  Optimizer & AdamW & AdamW \\
  Weight decay & 0.1 & 0.1 \\
  LR & \{0.1, 0.5, 1, 5\}$\times10^{-4}$ & \{0.1, 0.5, 1, 5\}$\times10^{-4}$ \\
  LR scheduler & Linear & Linear \\
  Warmup ratio & 10\% & 10\% \\
  Epochs & 20 & 10 \\
  Batch size & 32 & 32 \\
  \bottomrule
  \end{tabular}
  \centering
\end{table}

\subsection{Language modeling}
\label{sec: wiki details}
To study whether the QLLM can preserve the language modeling ability after finetuning, we finetune quantized Llama-2-7B, Llama-2-13B and Mistral-7B-v0.1 on the WikiText-2 \cite{wikitext} training set and report the perplexity on the validation set.

\textbf{Finetuning details.} The hyper-parameters for finetuning are listed in Table \ref{tab: hyper-parameter space for decoder-only}. We evaluate the finetuned QLLM on the validation set every epoch and report the best perplexity. After determining the best learning rate, we conduct three random runs and report the mean and standard deviation in Table \ref{tab: wiki and gsm8k}.

\begin{table}[ht]
  \centering
  \scriptsize
  \caption{Hyper-parameter search space for the finetuning of Llama-2 and Mistral. Please refer to Table \ref{tab: best hyper-paramete for wiki and gsm8k} for the best learning rate for different LLMs on WikiText-2 and GSM8K. \label{tab: hyper-parameter space for decoder-only}}
  \begin{tabular}{l|cc|cc}
  \toprule
  \textbf{Hyper-parameter} & \textbf{WikiText-2} & \textbf{GSM8K} & \textbf{Arithmetic reasoning}  & \textbf{Commonsense reasoning} \\
  \midrule
  Optimizer & \multicolumn{2}{c|}{AdamW} & \multicolumn{2}{c}{AdamW} \\
  Weight decay & \multicolumn{2}{c|}{0.1} & \multicolumn{2}{c}{1.0} \\
  LR & \multicolumn{2}{c|}{\{0.1, 0.5, 0.7, 1, 3, 4\}$\times 10^{-4}$} & \multicolumn{2}{c}{$3\times10^{-4}$} \\
  LR scheduler & \multicolumn{2}{c|}{cosine} & \multicolumn{2}{c}{linear} \\
  Warmup ratio & \multicolumn{2}{c|}{3\%} & \multicolumn{2}{c}{10\%} \\
  Epochs & 3 & 6 & \multicolumn{2}{c}{3} \\
  Batch size & 64 & 16 & \multicolumn{2}{c}{16}  \\
  Max sequence length & 1024 & 512 & \multicolumn{2}{c}{512} \\
  \bottomrule
  \end{tabular}
  \centering
\end{table}

\begin{table}[ht]
  \centering
  \scriptsize
  \caption{Best learning rate for different LLMs on the WikiText-2 and GSM8K tasks. \label{tab: best hyper-paramete for wiki and gsm8k}}
  \begin{tabular}{ll|lll|lll|lll}
  \toprule
  & & \multicolumn{3}{c|}{\textbf{Llama-2-7B}} & \multicolumn{3}{c|}{\textbf{Llama-2-13B}} & \multicolumn{3}{c}{\textbf{Mistral-7B-v0.1}} \\
  \textbf{Task} & \textbf{Method} & \textbf{4 Bits} & \textbf{3 Bits} & \textbf{2 Bits} & \textbf{4 Bits} & \textbf{3 Bits} & \textbf{2 Bits} & \textbf{4 Bits} & \textbf{3 Bits} & \textbf{2 Bits} \\
  \midrule
  WikiText-2 & ApiQ-lw & 4e-4 & 3e-4 & 4e-4 & 3e-4 & 3e-4 & 3e-4 & 1e-4 & 7e-5 & 7e-5 \\
  & ApiQ-bw & 3e-4 & 3e-4 & 3e-4 & 3e-4 & 3e-4 & 3e-4 & 1e-4 & 1e-4 & 1e-4 \\
  \midrule
  GSM8K & ApiQ-lw & 3e-4 & 3e-4 & 3e-4 & 4e-4 & 4e-4 & 3e-4 & 7e-5 & 7e-5 & 7e-5 \\
  & ApiQ-bw & 4e-4 & 4e-4 & 4e-4 & 3e-4 & 3e-4 & 4e-4 & 7e-5 & 7e-5 & 7e-5 \\
  \bottomrule
  \end{tabular}
  \centering
\end{table}

\subsection{Arithmetic reasoning (single-task)}
\label{sec: arithmetic details single task}
To study the arithmetic reasoning ability of QLLMs, we finetune quantized Llama-2-7B, Llama-2-13B and Mistral-7B-v0.1 on the GSM8K \citep{gsm8k} training set and report the accuracy on the test set. 

\textbf{Finetuning details.} The hyper-parameters for finetuning are listed in Table \ref{tab: hyper-parameter space for decoder-only}. We evaluate the finetuned QLLM on the test set every epoch and report the best accuracy. After determining the best learning rate, we conduct three random runs and report the mean and standard deviation in Table \ref{tab: wiki and gsm8k}.

\subsection{Arithmetic reasoning}
\label{sec: arithmetic details}
The setting here contrasts with the previous experiments where each task involves finetuning a separate QLLM. Instead, we adopt a unified strategy by finetuning a single QLLM across all tasks as delineated in \citet{llmadapter}. We finetune Llama-2-7B and Llama-2-13B on Math10K \citep{llmadapter} which is constructed from the training sets of GSM8K \citep{gsm8k}, MAWPS, MAWPS-single \citep{mawps} and AQuA \citep{aqua}. Then we evaluate the finetuned QLLM on the test sets of AQuA, GSM8K, MAWPS and SVAMP \citep{svamp}. Such a setting is more practical as LLM is frequently used as a general model for various tasks.

\textbf{Finetuning details.} We follow \citet{llmadapter} to choose the hyper-parameters as in Table \ref{tab: hyper-parameter space for decoder-only}. Instead of evaluating the finetuned QLLM every epoch, we only evaluate the trained model from the last epoch, simulating the practical finetuning scenario. We conduct three random runs and report the mean and standard deviation in Table \ref{tab: arithmetic reasoning}. 

\subsection{Commonsense reasoning}
\label{sec: commonsense details}
In assessing the capacity of QLLM for commonsense reasoning, we focus on eight representative tasks: BoolQ \citep{boolq}, PIQA \citep{piqa}, SIQA \citep{siqa}, HellaSwag \citep{hellaswag}, WinoGrande \citep{winogrande}, ARC-e, ARC-c \citep{arc}, and OBQA \citep{obqa}. Similar to the multiple arithmetic reasoning tasks, we follow the setting of \citet{llmadapter} and finetune a single QLLM across all tasks. Specifically, the training and test sets from these eight tasks are reformulated according to a predefined template, so all tasks can be trained or evaluated in a generative way. Then we finetune Llama-2-7B and Llama-2-13B on the combined training set and report the accuracy on the test sets.

\textbf{Finetuning details.} We also borrow the finetuning recipe of \citet{llmadapter} as in Table \ref{tab: hyper-parameter space for decoder-only}. We only evaluate the trained model from the last epoch (simulate a practical finetuning scenario), conduct three random runs, and report the mean and standard deviation in Table \ref{tab: commonsense reasoning results}.

\begin{figure}[t]
  \includegraphics[width=0.98\linewidth]{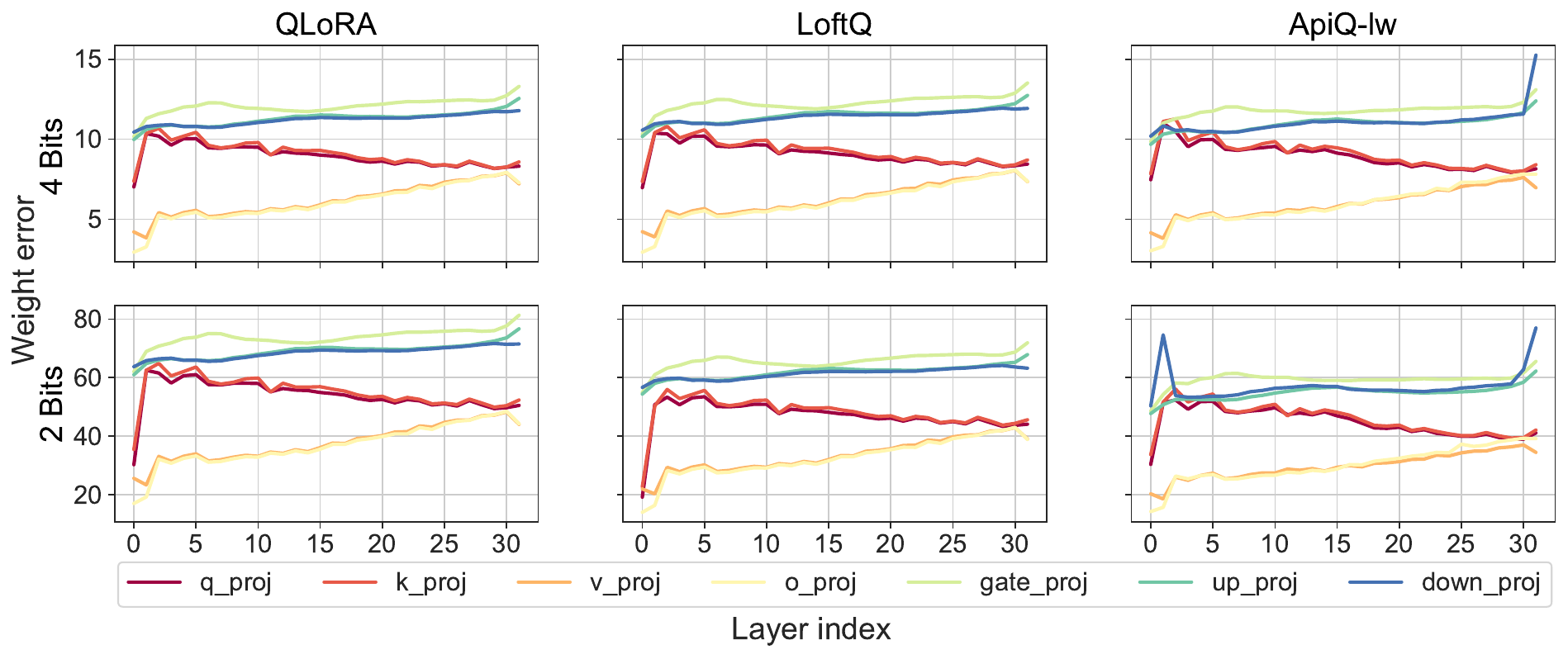}
  \caption{The weight quantization error $||\bm{W} - (\bm{Q} + \bm{AB}^T)||_F$ for different linear layers of Llama-2-7B. For 4-bit quantization, all methods are comparable, because 4-bit quantization doesn't significantly break down the starting point. For 2-bit quantization, ApiQ has the smallest quantization error for most layers, although its goal is to minimize the activation error.}
  \label{fig: weight diff}
\end{figure}

\begin{figure}[ht]
 \centering
  \begin{subfigure}
    \centering\includegraphics[width=0.49\columnwidth]{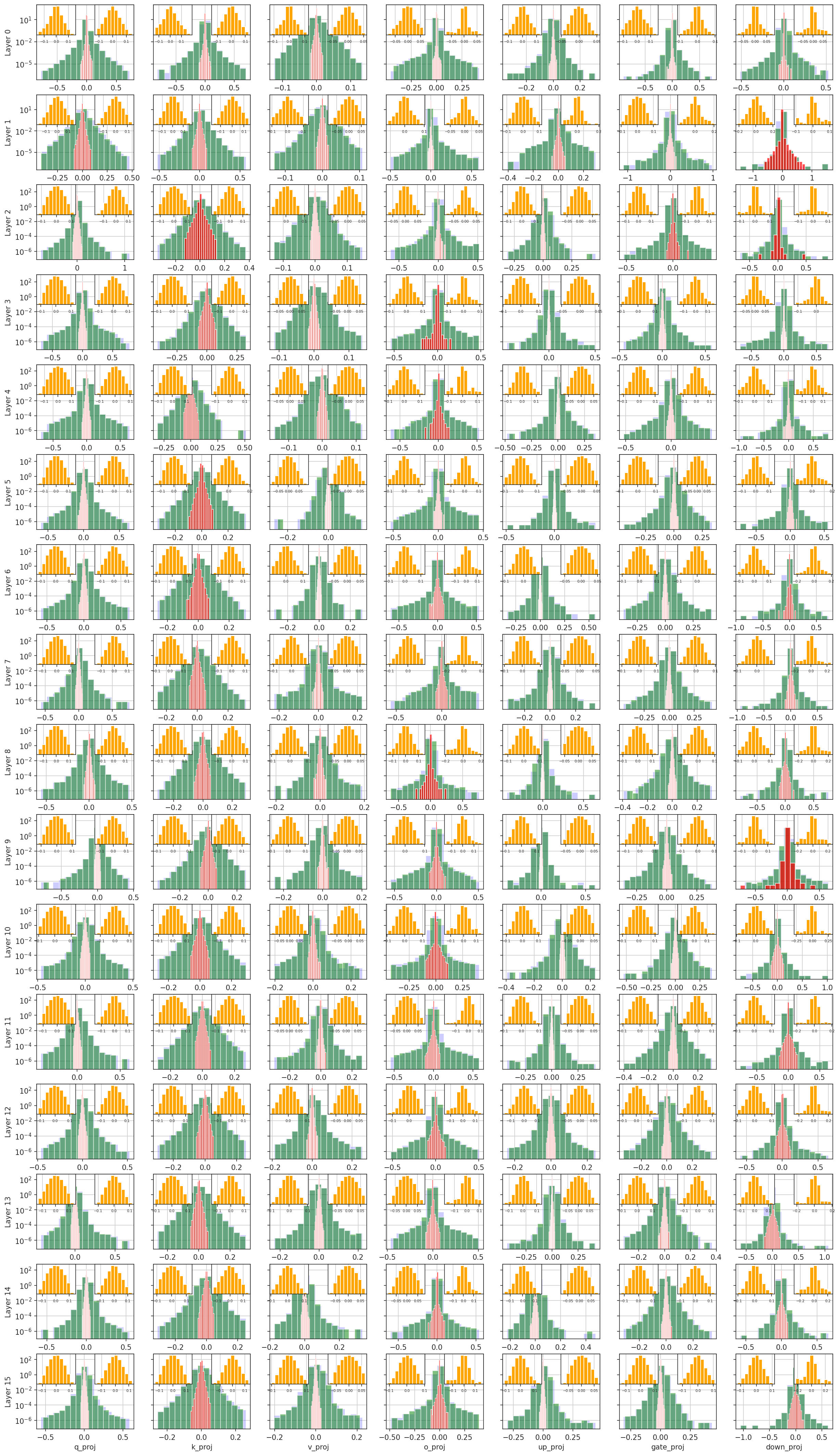}
  \end{subfigure}
  \hfill
  \begin{subfigure}
    \centering\includegraphics[width=0.49\columnwidth]{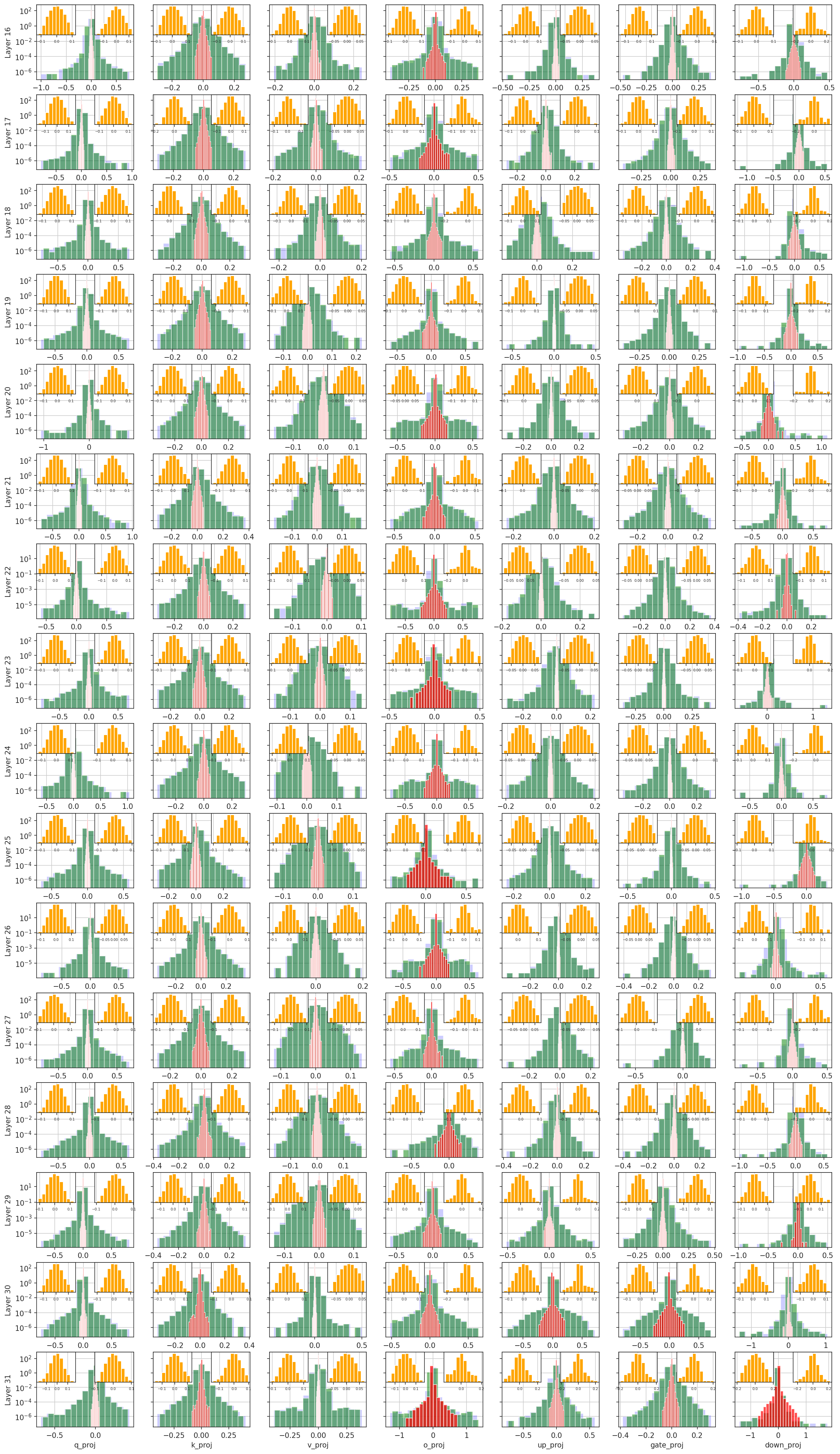}
  \end{subfigure}
  \caption{Histogram of $\bm{Q}$, $\bm{A}$ and $\bm{B}$ for the \textbf{4-bit} quantized layer of Llama-2-7B with \textbf{ApiQ-lw}. Blue: \textcolor{blue}{$\bm{W}$}. Green: \textcolor{green}{$\bm{Q}$}. Red: \textcolor{red}{$\bm{AB}^\top$}. Orange: \textcolor{orange}{$\bm{A}$}(Left) or \textcolor{orange}{$\bm{B}$}(Right). Compared to LoftQ, the distribution of $\bm{B}$ of ApiQ is symmetric and doesn't have outliers, which might be one reason why ApiQ outperforms LoftQ.\label{fig: hist 4bit}}
\end{figure}

\begin{figure}[ht]
  \centering
  \begin{subfigure}
    \centering\includegraphics[width=0.49\columnwidth]{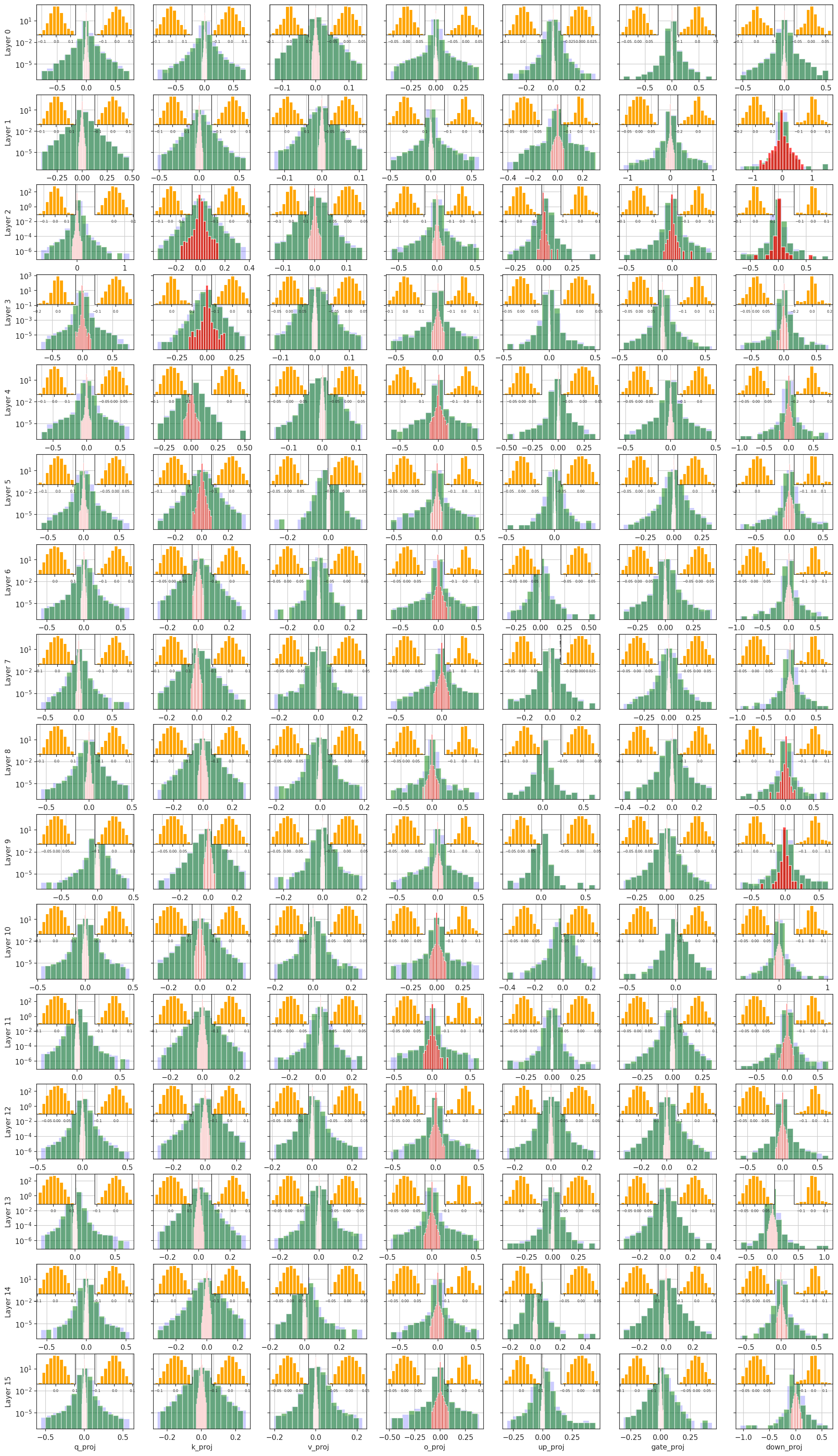}
  \end{subfigure}
  \hfill
  \begin{subfigure}
    \centering\includegraphics[width=0.49\columnwidth]{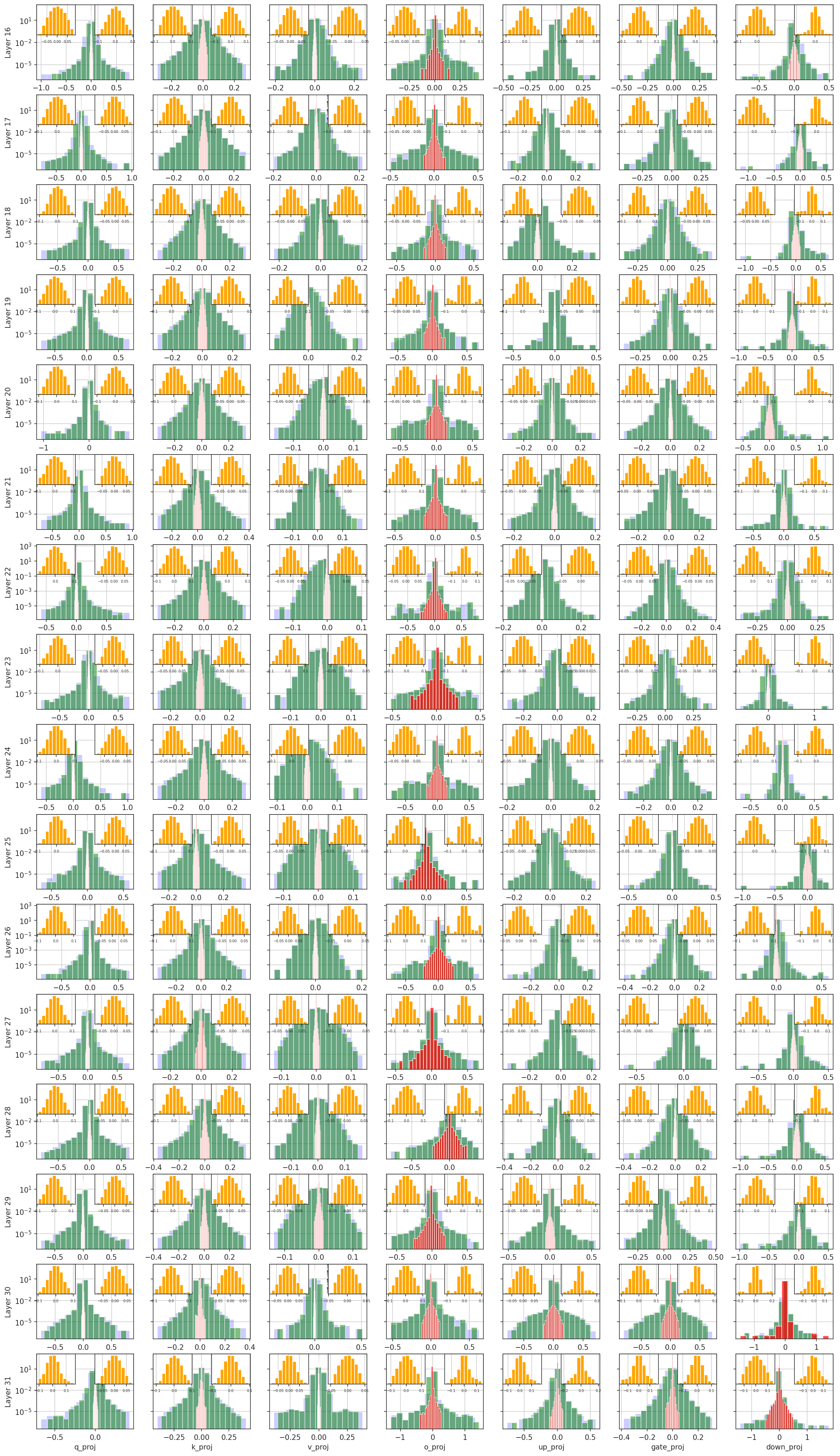}
  \end{subfigure}
  \caption{Histogram of $\bm{Q}$, $\bm{A}$ and $\bm{B}$ for the \textbf{3-bit} quantized layer of Llama-2-7B with \textbf{ApiQ-lw}. Blue: \textcolor{blue}{$\bm{W}$}. Green: \textcolor{green}{$\bm{Q}$}. Red: \textcolor{red}{$\bm{AB}^\top$}. Orange: \textcolor{orange}{$\bm{A}$}(Left) or \textcolor{orange}{$\bm{B}$}(Right). Compared to LoftQ, the distribution of $\bm{B}$ of ApiQ is symmetric and doesn't have outliers, which might be one reason why ApiQ outperforms LoftQ.\label{fig: hist 3bit}}
\end{figure}

\begin{figure}[ht]
  \centering
  \begin{subfigure}
    \centering\includegraphics[width=0.49\columnwidth]{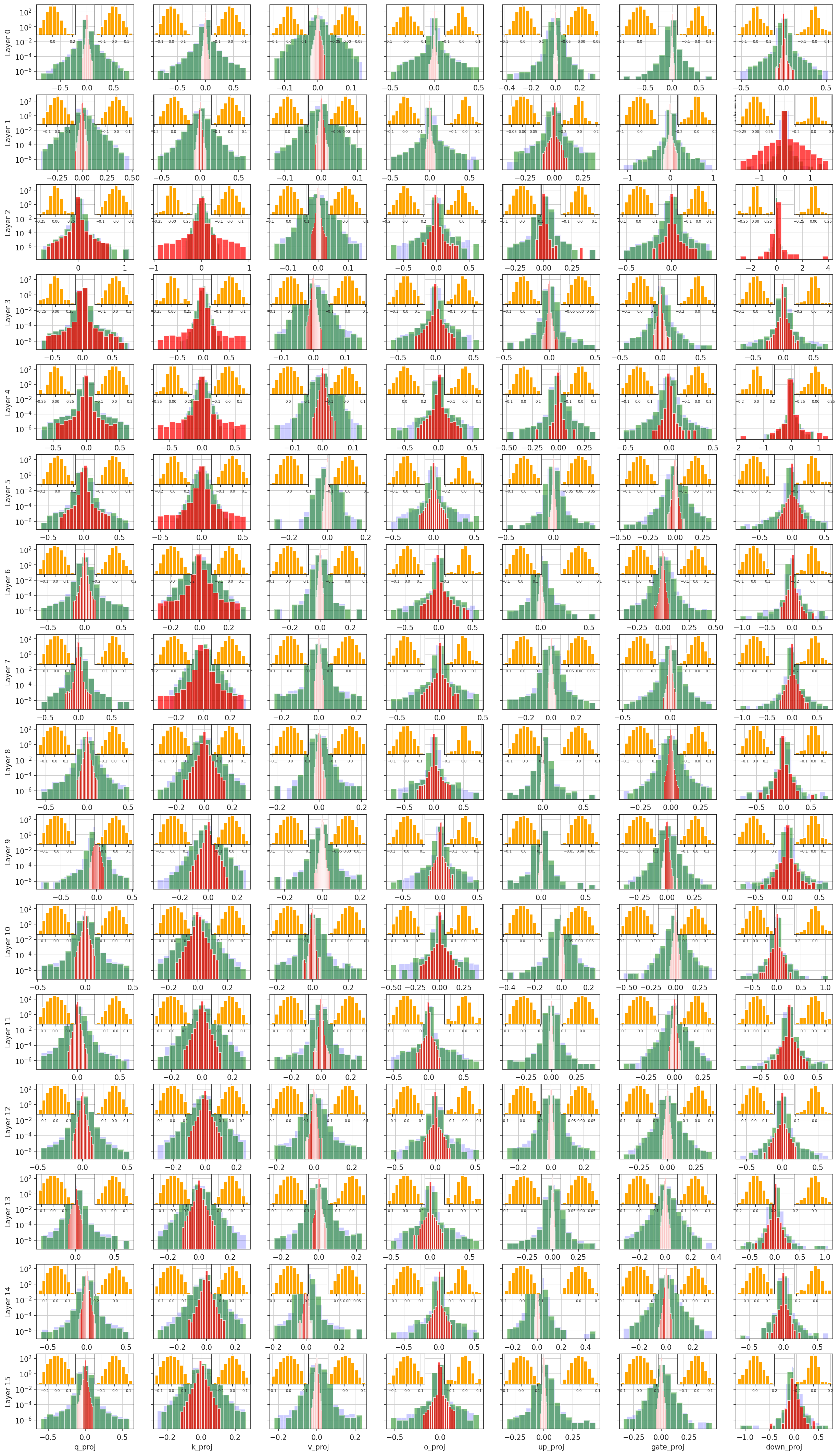}
  \end{subfigure}
  \hfill
  \begin{subfigure}
    \centering\includegraphics[width=0.49\columnwidth]{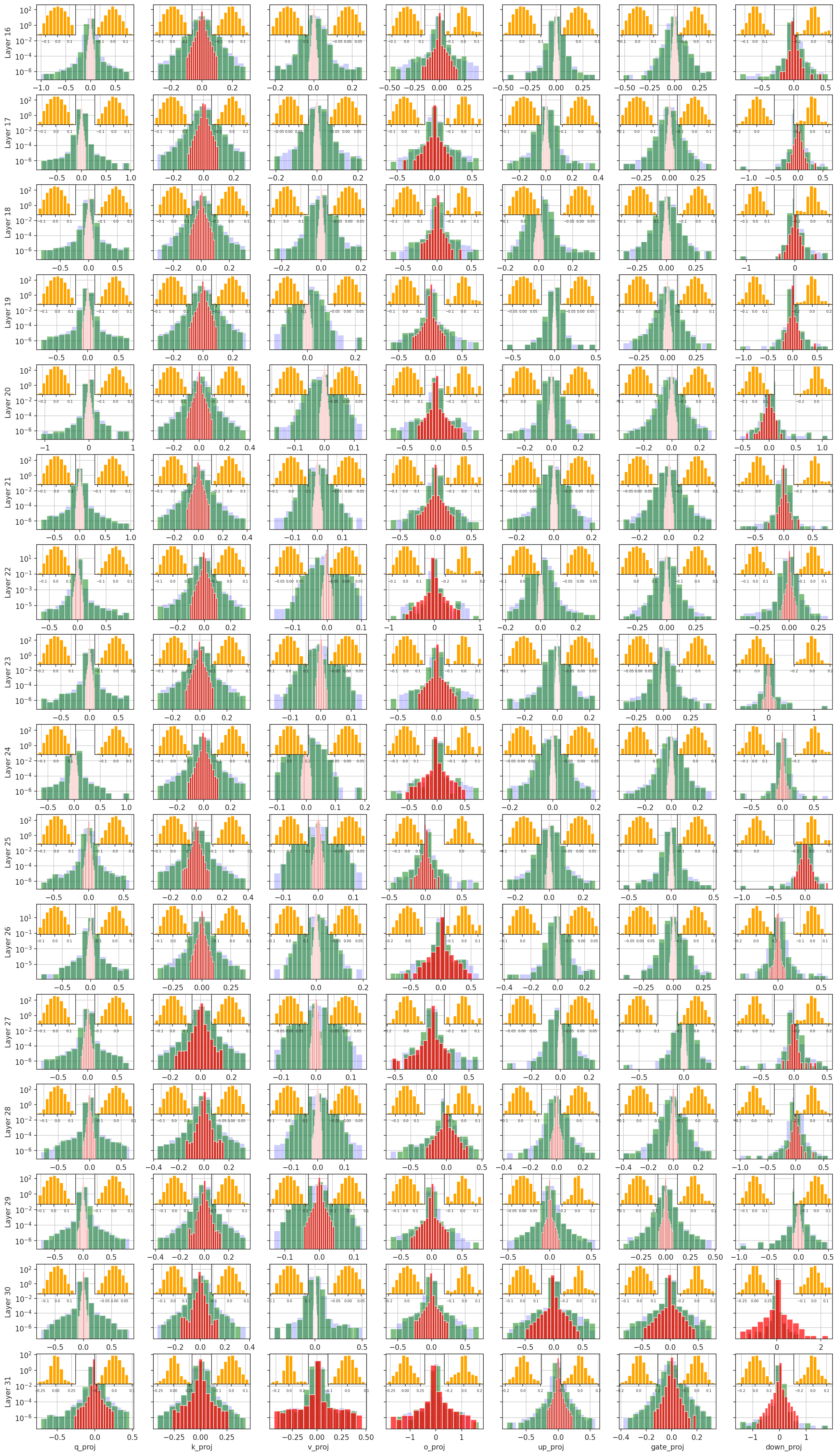}
  \end{subfigure}
  \caption{Histogram of $\bm{Q}$, $\bm{A}$ and $\bm{B}$ for the \textbf{2-bit} quantized layer of Llama-2-7B with \textbf{ApiQ-lw}. Blue: \textcolor{blue}{$\bm{W}$}. Green: \textcolor{green}{$\bm{Q}$}. Red: \textcolor{red}{$\bm{AB}^\top$}. Orange: \textcolor{orange}{$\bm{A}$}(Left) or \textcolor{orange}{$\bm{B}$}(Right). Compared to LoftQ, the distribution of $\bm{B}$ of ApiQ is symmetric and doesn't have outliers, which might be one reason why ApiQ outperforms LoftQ.\label{fig: hist 2bit}}
\end{figure}

\begin{figure}[ht]
  \centering
  \begin{subfigure}
    \centering\includegraphics[width=0.49\columnwidth]{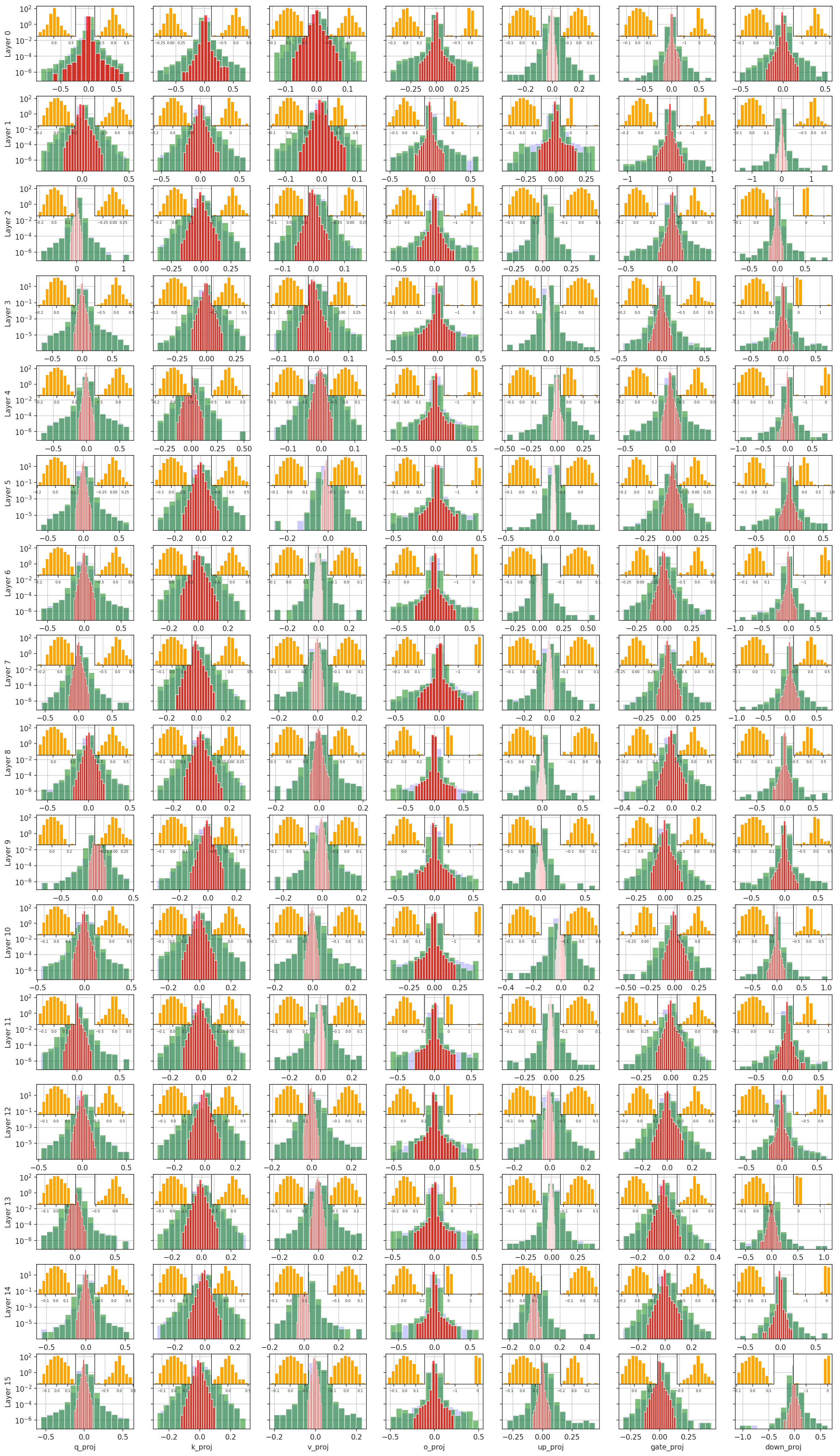}
  \end{subfigure}
  \hfill
  \begin{subfigure}
    \centering\includegraphics[width=0.49\columnwidth]{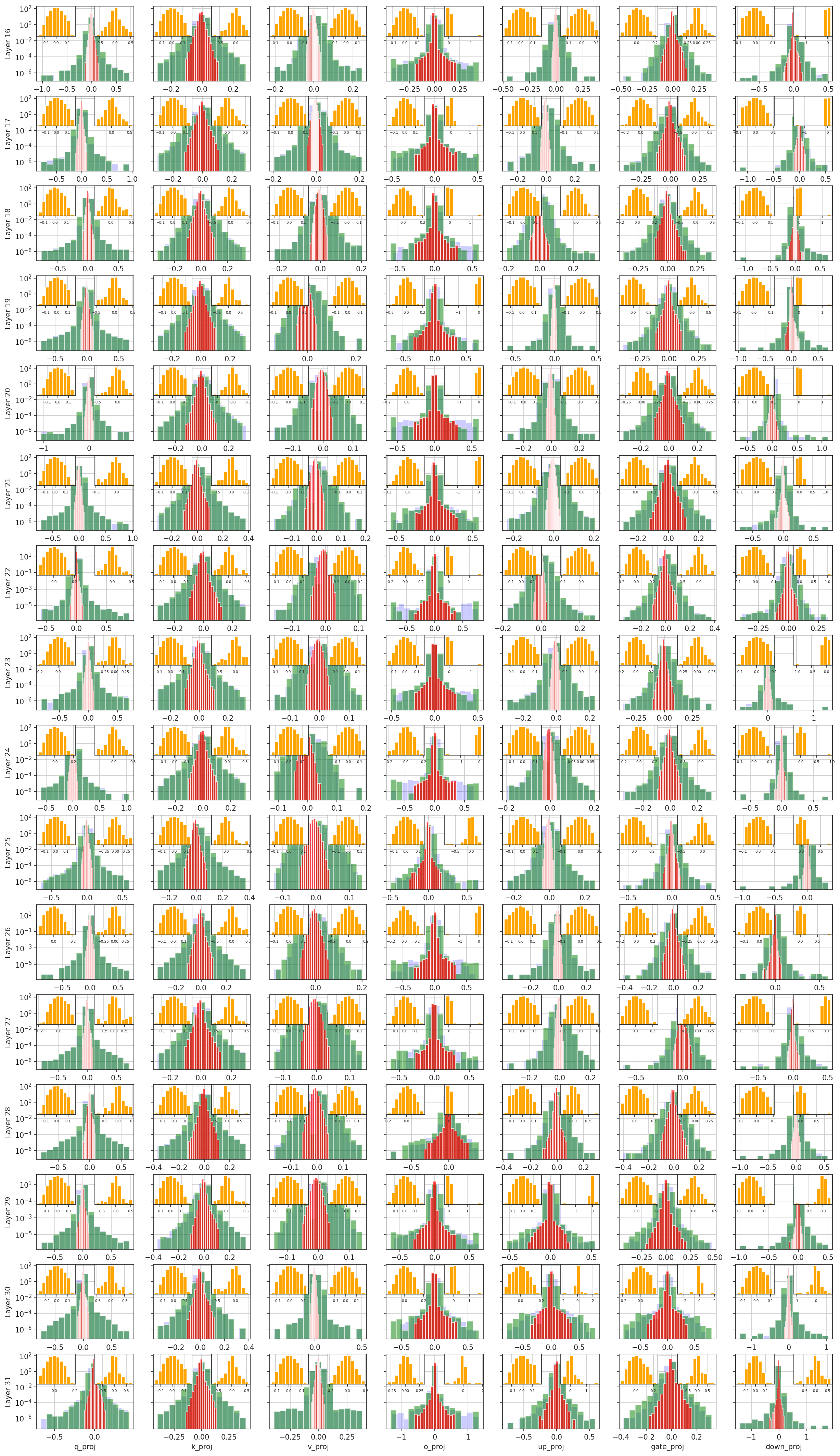}
  \end{subfigure}
  \caption{Histogram of $\bm{Q}$, $\bm{A}$ and $\bm{B}$ for the \textbf{2-bit} quantized layer of Llama-2-7B with \textbf{LoftQ}. Blue: \textcolor{blue}{$\bm{W}$}. Green: \textcolor{green}{$\bm{Q}$}. Red: \textcolor{red}{$\bm{AB}^\top$}. Orange: \textcolor{orange}{$\bm{A}$}(Left) or \textcolor{orange}{$\bm{B}$}(Right). Compared to ApiQ, the distribution of $\bm{B}$ of LoftQ is asymmetric for most linear layers and has many outliers, which might be one reason why LoftQ performs worse for 2-bit quantization. \label{fig: hist loftq 2bit}}
\label{icml-historical}
\end{figure}


\end{document}